\documentclass[review]{elsarticle}

\usepackage{setspace} 
\pdfoutput=1
\usepackage{graphicx}
\usepackage{caption}
\usepackage{subcaption}
\usepackage{fullpage}
\usepackage{amssymb}
\usepackage{amsmath}
\usepackage{color}
\usepackage{amsmath}
\usepackage{algpseudocode}
\usepackage{algorithm}
\usepackage[titletoc,title]{appendix}
\usepackage{epstopdf}

\journal{}

\biboptions{authoryear}

\usepackage[table,xcdraw]{xcolor}
\usepackage{booktabs}
\usepackage[para,online,flushleft]{threeparttable}
\usepackage{color}
\usepackage[table,xcdraw]{xcolor}
\definecolor{light-gray}{gray}{0.85}

\begin{document}
\begin{frontmatter}

\title{The detour problem in a stochastic environment: Tolman revisited}

\author{Pegah Fakhari\fnref{mycorrespondingauthor}}
\author{Arash Khodadadi}
\author{Jerome R. Busemeyer}
\address{Indiana University, Department of Psychological and Brain Sciences, Bloomington, IN, United States}
\fntext[mycorrespondingauthor]{Corresponding author. Indiana University, Department of Psychological and Brain Sciences, 1101 E. 10th street, 47405-7007, Bloomington, IN, United States.}

\begin{abstract}
We designed a grid world task to study human planning and re-planning behavior in an unknown stochastic environment. In our grid world, participants were asked to travel from a random starting point to a random goal position while maximizing their reward. Because they were not familiar with the environment, they needed to learn its characteristics from experience to plan optimally. Later in the task, we randomly blocked the optimal path to investigate whether and how people adjust their original plans to find a detour. To this end, we developed and compared $12$ different models. These models were different on how they learned and represented the environment and how they planned to catch the goal. The majority of our participants were able to plan optimally. We also showed that people were capable of revising their plans when an unexpected event occurred. The result from the model comparison showed that the model-based reinforcement learning approach provided the best account for the data and outperformed heuristics in explaining the behavioral data in the re-planning trials.  
\end{abstract}
\end{frontmatter}

\section{Introduction}
Humans deal with planning problems in their everyday situations. One of the very familiar situations is to navigate from one place to another in a neighborhood or city. In this scenario, usually there is more than one path to choose, and, depending on the goal, one might select the shortest path, the city roads, a bypass/highway outside the traffic area or a path with the minimal traffic lights. Once he chooses the highway, he still needs to decide whether to take the toll line and/or where to exit. On the other hand, if he chooses the city roads, he would need to decide which intersection to go, to use the main street or to use shortcuts, etc. In other words, after selecting a general path (plan), there are still small paths decisions. 

This is an example of a more general problem in which one needs to optimally plan a sequence of interdependent choices to accomplish a goal. In some cases, the shortest path is the optimal path and in others his goal might be to avoid the traffic at all costs.

\begin{figure}
	\includegraphics[scale=.5]{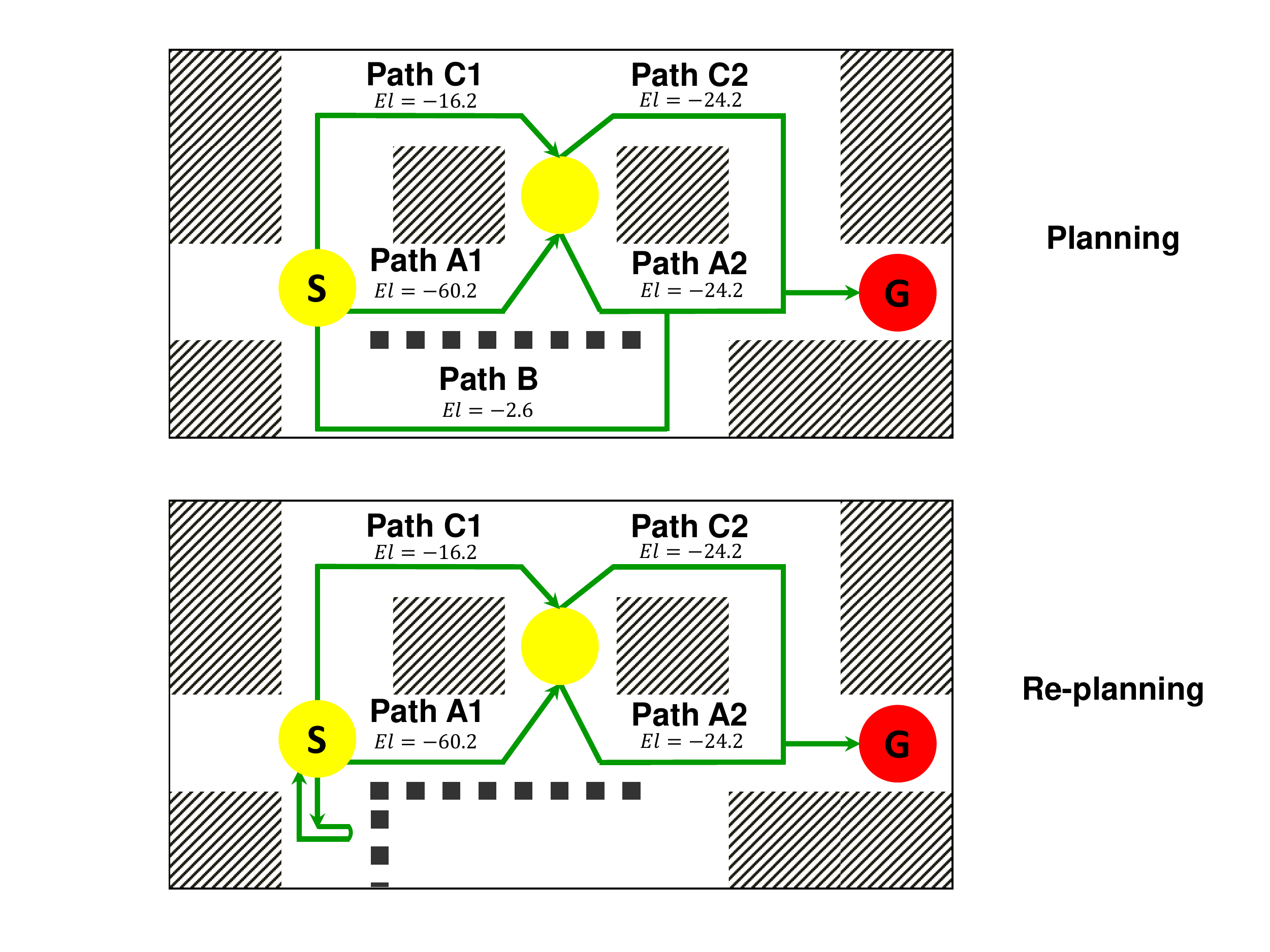}
	
	\caption{{\bf General grid world used in our experiments. The decision nodes are represented by yellow circles and the goal is depicted by red circle. The green paths are available to participants. Note that participants can not see the obstacles depicted by shadow areas. The black dashed line in the grid separates path B and path A1A2. Top: In the planning trials, for this current start and goal positions, $5$ paths are available to participants and they need to find the optimal path. Down: In the re-planning trials, the optimal path, path \textbf{B} is blocked and participants need to find the detour.}} \label{GenF}
\end{figure}

In multistage decision making, unlike isolated choices, the focus is on how people analyze the interrelated choices to make an \textit{optimal} sequence of choices, \cite{hotaling_dynamic_2015}, \cite{gonzalez_dynamic_2017} \footnote{Reference style was generated by the latex template (i.e. APA format). Due to our previous experience, it will be fixed before publication}. Usually, sequential decisions are represented in a decision tree in which the result of an action at one stage (e.g. a decision node) will be fed into the next stage, which might be another decision node or possibly an output node. Consider a decision tree with two decision nodes (yellow circles) and five possible (green) paths that is represented in the upper panel of Fig.~\ref{GenF}. Given the starting position, \textbf{S} and the goal, \textbf{G}, the paths are Path B, Path A1A2, Path A1C2, Path C1C2 and Path C1A2. The black dashed line in the grid separates path B and path A1A2. 

In order to make optimal choices, one should know the actual output of each decision node and the uncertainty of each transition. For instance, although the number of steps (or actions) between the starting position and the next decision node (or the goal) is not depicted in Fig.~\ref{GenF}, based on the expected losses ($El$), we know that path B is the best path to go to the goal position. In an experience-based decision tree, this knowledge is established from (an individual's) experience, \cite{daw_model-based_2011}, \cite{momennejad_successor_2016}, \cite{huys_bonsai_2012}, \cite{dezfouli_habits_2012} and \cite{keramati_adaptive_2016} while in a description-based version, it is provided by the experimenter (available during the task), \cite{hotaling_dft-d:_2012}, \cite{hey_strategies_2011}, \cite{dorner_errors_1994}, \cite{johnson_multiple-stage_2001}, \cite{johnson_dynamic_2005}. However, having this knowledge cannot guarantee the optimal behavior, \cite{hey_people_2005}, \cite{sims_melioration_2013}, \cite{yechiam_melioration_2003}, \cite{hotaling_dft-d:_2012}, \cite{huys_interplay_2015}, \cite{momennejad_successor_2016}, \cite{huys_interplay_2015}, \cite{botvinick_hierarchically_2009}, \cite{keramati_speed/accuracy_2011}.

\subsection{Examining the Optimality of re-planning in sequential decision making tasks}
In many of the previous studies on planning in decision trees, the environment is not dynamically changing. Once the participant learns about the risky sequential decision environment, she can \textit{optimally} plan her actions and does not need to update her knowledge later. But in real life, our environment is always changing and unexpected events happen. Usually, we have two strategies to deal with these situations: reevaluate our plans with the new information (which is also known as re-planning) or ignore the new information and stick to our original plan. 

In this article, we extend previous planning experimental designs to situations in which the participants experience random changes in the environment and need to modify their original plans to get to the goal position. We test learning to plan and re-planning in one unique framework: a $4$ by $7$ grid world with stochastic losses. In the learning phase of our experiments, we look into the planning behavior and how participants can learn to find the optimal sequence of choices. Then, in the test phase, we block the optimal path randomly and ask our participants to find a detour path (re-planning behavior) based on what they have learned during the learning phase as illustrated in the bottom panel of Fig.~\ref{GenF}. On $30\%$ of the trials, path B (the \textit{optimal} path) is randomly blocked and becomes unavailable \footnote{Participants do not experience the blockage unless they check the optimal path.}. The vertical black dashed line that is added to the grid shows the wall that makes path B unavailable. In this situation, path C1A2 is the best path to choose (optimal re-planning behavior). 

It is important to emphasize that the starting and goal positions are not fixed in our design. We randomize these pairs for three reasons: 1) To make sure that our participants have a fair exposure to different aspects of the grid world environment; 2) A random starting point and goal makes it more challenging to discriminate different models and their predictions; 3) To examine human planning behavior from different arbitrary decision nodes located in different layers of a decision tree.

In section ~\ref{ModelFit}, we discuss how we fit different models and compare the results in detail. However, it is important to highlight that our design includes a generalization test that fits model parameters to the planning phase (with no blocks in the optimal path), and then subsequently uses these same parameters to predict re-planning in a generalization test when blocks are introduced, \cite{busemeyer_model_2000}. This provides a very strong test of the competing models that vary in number of parameter and model complexity. In other words, our model comparison is not restricted to how different models can learn the model of the environment and whether or not they can predict planning (which they have been trained for), but involves a more rigorous test on how they can perform in an unexperienced environment. \\

\subsection{Computational models for learning sequential decision making from experience}
In most everyday situations, it is almost impossible to have access to all of the information required to make sequential decisions from the start, and much of this information must be learned from experience. While it is necessary to take advantage of previous experiences to make better choices in the future, it is also essential to evaluate actions and assign (proper) credit to the earlier choices as well as to the later actions in a sequence (temporal credit assignment). These two criteria are the heart of many reinforcement learning (RL) and unsupervised learning algorithms in computer science \cite{walsh_navigating_2014}, \cite{sutton_reinforcement_1998}, \cite{bertsekas_neuro-dynamic_1996}. 

In RL, the goal is to choose appropriate action(s) which maximize the expected sum of future rewards. In this set of problems, the agent initially does not know the correct correspondence between states and actions and receives feedback following an action or a sequence of actions. This framework can address learning mechanisms, ranging from the very basic stimulus response (habitual behavior) to the more complicated goal-directed behavior (including, but not limited to, planning). Whether people conduct themselves in accordance with the RL predictions in sequential choices is still under investigation and debate (for a detailed review, see \cite{walsh_navigating_2014}). For instance, \cite{huys_bonsai_2012}, asked their participants to plan a sequence of $2$-$8$ choices by traveling through $6$ states (represented as boxes on the screen) to maximize their score. Different RL models were fitted to explain the underlying mechanism. They found that people extensively pruned sub-trees with large losses in order to reduce the size of the decision problem into a computationally manageable problem \cite{huys_bonsai_2012}\footnote{They used two different discounting parameters to differentiate the greatest loss from the other losses.}.  

In general, there are two classes of RL models that can provide a solution to our sequential decision problem. The first class finds the optimal policy by learning the model of the environment and is called the model-based approach. The second class, the model-free approach, is able to maximize the expected sum of future rewards \textit{without} knowing the characteristics of the environment, \cite{sutton_reinforcement_1998}, \cite{daw_uncertainty-based_2005}. Therefore, the second approach is computationally efficient. But as we explain later, since it only incorporates one-step rewards (plus the expected future reward from the next step) from a particular state into planning which limits its ability to plan when the start and goal positions change.

There is a third RL algorithm, called successor representation, that learns a rough representation of the environment by storing the expected future visits of each state, \cite{gershman_successor_2012}, \cite{dayan_improving_1993}. It is computationally less expensive than the model-based RL and can easily explain the planning behavior in our experiments it has limited ability to re-plan in a changing environment.

Last but not least, we investigate how heuristics and simple strategies perform in our risky sequential choice environment. In section ~\ref{Heuristic}, we show the predictions of $3$ different heuristic based models. Our design was able to show that these heuristic models fail, both quantitatively and qualitatively.\\

\section{Material and Methods}  \label{MatAndMeth}
\subsection{Overview}
Learning the characteristics of the environment to plan optimally (and possibly re-plan in case of the unexpected changes) was first examined by Tolman and Honzik with rats, \cite{tolman_cognitive_1948}, \cite{tolman_introduction_1930}. In this task, which is known as the detour experiment, first, rats were exposed to three different paths with different lengths to a goal/food location similar to Fig.~\ref{MM01}, a. Then in the test phase, they blocked the shortest path and checked the rats' choice in finding the second shortest path \cite{voicu_latent_2002}. \cite{tolman_cognitive_1948} and \cite{tolman_introduction_1930} summarized that acquiring skills like turning right at specific positions to do a particular task (e.g. reaching a goal/food) could be explained by a stimulus-response learning system. However, only goal directed behavior could explain the rapid learning curve and fewer errors in rats' behavioral data in finding the detour paths, \cite{tolman_studies_1946}. 

Inspired by these findings, one of our very basic questions in this study is whether people can plan in a stochastic environment and if so how far they can modify their plans in order to maximize their earnings. There are two phases in our experiment: a training phase which allows participants to explore and learn about the grid world (without monetary reward) and a test phase that goes beyond optimal planning and requires finding the second best path to maximize their money. One of the main differences between our design and Tolman's detour problem is that in \cite{tolman_cognitive_1948}, the environment is deterministic but in our grid world experiment, there are multiple cells with stochastic rewards ($2$ cells in experiment $1$ and $5$ cells in experiments $2$ and $3$). Therefore, the \textit{best} path is not simply the \textit{shortest} path but is the \textit{optimal} path which requires planning. Recognizing the difference between the shortest and optimal paths plays a key role on the (amount of) money that participants can earn.

We should also note that in our design, we used random pairs in each trial to make sure that participants explore the entire grid world. However, in the original detour problem, Tolman \cite{tolman_cognitive_1948} exposed rats to different paths using fixed starting and goal positions. 

Table~\ref{tab:SummaryOfExp} highlights the differences and similarities among the three experiments. It is important to emphasize that participants can only see a plain grid on the screen along with their current position (yellow circle) and their destination (red circle, Fig.~\ref{MM01}) with no sign/cue of the obstacles or the stochastic losses. They do not have access to papers nor calculators/cellphones to do any computations or to take notes. In the instruction, they are told to consider a scenario that they move to a new city (a $4$ by $7$ grid) and need to get from one place to another (determined by the yellow and red circles). In the test phase, participants are told that a random accident might happen in one of the possible routes and block that path. If they see the accident they need to find a detour path.

   \begin{table}[ht]
   \caption{Summary Of Experiments} 
   \centering 
   \begin{tabular}{l c c c c} 
   \hline\hline 
   Design & Experiment $1$ & Experiment $2$ & Experiment $3$\\ [0.5ex] 
   \hline 
   Number of learning blocks & $6$ & $6$ & $6$ \\
   Number of test blocks & $2$ & $2$ & $3$\\
   Number of pretest blocks & $1$ & $1$ & $1$ \\
   Number of probabilistic losses & $2$ & $5$ & $5$ \\ 
  Fixed G and S in the test blocks & Yes & Yes & \textbf{No} \\ 
  Fixed G and S in the learning blocks & No & No & No \\
  Fixed G and S in the pretest block & No & No & No \\
  Cells with stochastic losses & $15$, $16$ & $9$, $11$, $16$, $19$, $17$ & $9$, $11$, $16$, $19$, $17$ \\
  Number of re-planning trials in the test trials & $13$ out of $40$ & $13$ out of $40$ & $20$ out of $60$ \\ [1ex] 
   \hline 
   \end{tabular}
   			\begin{tablenotes}
   				\footnotesize Experiment $1$ has two stochastic losses with one sure loss at cell $21$ with the loss of $-45$. In experiment $2$ and $3$, there are $5$ stochastic losses (similar environment). The rate of re-planning trials in the test blocks is fixed in all three experiments ($33 \%$). Unlike experiments $1$ and $2$, the pairs in the test blocks of the experiment $3$ are random. In all three experiments, there is one pretest block and six learning blocks. G and S stand for Goal and Starting positions.
   			\end{tablenotes}
   \label{tab:SummaryOfExp} 
   \end{table}
 
 In the following section, first we summarize the behavioral results for these three experiments and then we try to characterize participants' strategies in planning and re-planning in the pretest and the test blocks respectively. Presumably any candidate model must incorporate environmental changes in its planning approach in order to make sure that a new information is propagated to the action selection module promptly. 
 
\subsection{Experiment 1} 
\subsubsection{Participants}
Twenty healthy ($11$ females) participants performed experiment $1$ for payment. Minimum payment was $\$9$ and participants could earn up to $\$16$ based on their performance during the task. Informed consent was collected from all participants and the study was approved by Indiana University Institutional Review Boards. 
   
\subsubsection{Task} \label{Exp1}
\begin{figure}
\includegraphics[scale=.5]{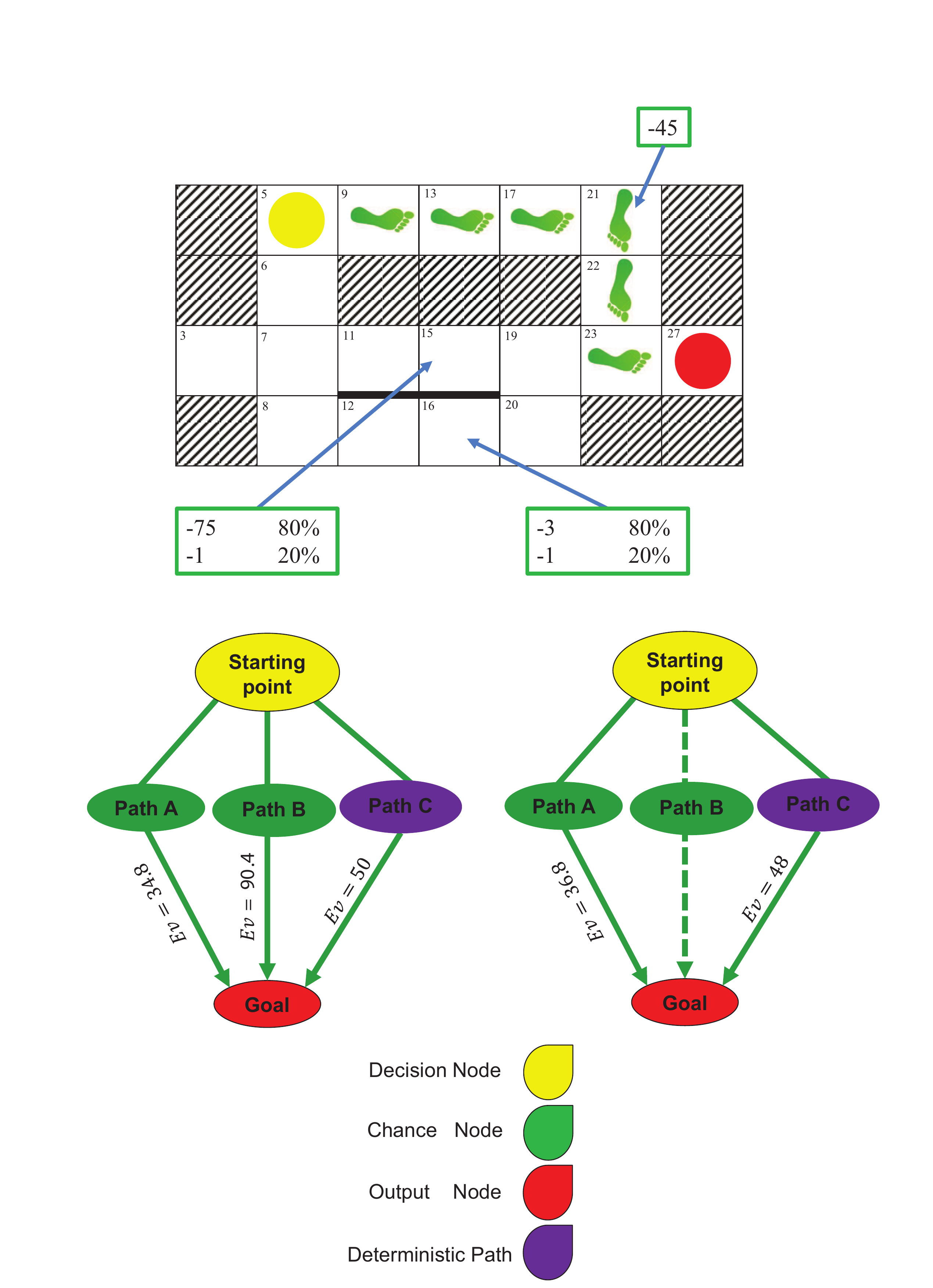}
    
\caption{{\bf Experiment $1$. Up: A $4$ by $7$ grid world experiment. Green footsteps indicate one possible path from starting point to the goal (path C). Expected values ($Ev$) are calculated based on the current starting point. Down: Corresponding decision tree for planning (left) and re-planning (right). There are $3$ possible paths between the starting position and the goal. Expected value in the re-planning phase is different from planning phase and is calculated from cell $7$. Cell $7$ is the position that participants choose their detour path in the re-planning trials.}} \label{MM01}
\end{figure}

Participants learned to \textit{correctly} travel through a grid world in as few moves as possible while maximizing their reward. In experiment $1$, there were $9$ blocks and each has $20$ trials. In each trial participants viewed a typical $4$ by $7$ grid world example and were asked to start from a \textit{random} starting point, represented by a yellow circle, and reach to a random goal position (in red circle) using arrow keys (up, right, down and left), as shown Fig.~\ref{MM01}, a. For instance in Fig.~\ref{MM01}, the start and goal points are arbitrarily located at cell $5$ (start) and cell $27$ (goal). 

Except for the goal point, participants received either a fixed or stochastic “punishment” for any movement in the grid world. If they caught the goal, they earned $+100$ points (later exchanged for money at an exchange rate equal to $0.01$). The fixed \textit{regular} punishment was delivered in transition to most of the cells and it was equal to $-1$. At cell $21$, there was another deterministic (but not regular) punishment. If participants entered this cell, they received $-45$. Finally, cells $15$ and $16$ had stochastic punishments. If transmitted to these cells, participants received a different punishment ($-75$, $-3$ respectively) with probability of $0.8$ and a regular punishment ($-1$) with probability of $0.2$. It has been suggested by \cite{hertwig_decisions_2004} that the existence of rare events, specially outcomes that occur with probability less than $0.15$, can lead to non-optimal behavior. Using this current settings for the loss structure, we have one optimal, one sub-optimal and one non-optimal path (three distinct paths). For instance, starting at cell $3$, path A through $3$, $7$, $11$, $15$, $19$, $23$, $27$ is the non-optimal path. Path B through $3$, $7$, $8$, $12$, $16$, $20$, $19$, $23$, $27$ and path C through $3$, $7$, $6$, $5$, $9$, $13$, $17$, $21$, $22$, $23$, $27$ are the optimal and sub-optimal paths. Note that the probabilities, the amount and the positions of the punishments were not known to participants.

In each attempt to catch the goal, a participant had $15$ moves. There were ten hidden fixed obstacles which prevented participants to go to the selected direction while traveling in the grid world (e.g. selecting up at cell $11$ was blocked). Hidden obstacles are shown with upward diagonal texture in Fig.~\ref{MM01}, a. Each time a participant hit an obstacle, she received a punishment of $-1$ and used up one of her $15$ moves. The first six blocks were designed to help participants learn about the obstacles, punishments and the details of the grid world (learning phase). During this phase participants were not getting paid. 

The $7$th block was designed to check optimal planning. Similar to previous blocks, it consisted of $20$ random pairs of starting and goal points but the difference was that participants got paid based on their score (we called it the pretest block). Finally in the last two blocks ($8$th and $9$th), optimal path, Path B, was blocked in one-third of the trials. The random blockage was to further examine optimal planning and also re-planning (finding the second optimal path) in our dynamic environment. The starting and goal positions were fixed at cell $3$ and cell $27$ in $8$th and $9$th blocks (the test blocks).
  
In Fig.~\ref{MM01}, bottom panel, we present two decision trees. One on the left, all three paths are available (planning phase) and their expected values were calculated based on the current positions of the start (cell $5$) and goal (cell $27$). For instance if one chooses path A (through cell $6$, $7$, $11$, $15$, $19$, $23$, $27$), the expected value is $34.2$, if he chooses path B (through $6$, $7$, $8$, $12$, $16$, $20$, $19$, $23$, $27$), the expected value is $90.4$ and finally for path C (through $9$, $13$, $17$, $21$, $22$, $23$, $27$), the exact value is $50$ since all punishments are deterministic. The decision tree on the right is related to re-planning phase where path B is not available (dashed line). The expected values are calculated from cell $7$ because this cell is the starting point to find the detour path after the participant experiences the random blockage in the optimal path (path B) at cell $8$. Then he goes up to cell $7$ to choose one of the available paths (path A and path C). If he chooses path A, then the expected value is $38.8$ while choosing path C leads to the value of $48$. 

Note that in experiment $1$, the corresponding decision trees in planning are directly related to the grid world on the top panel of Fig.~\ref{MM01} and where the starting point and goal are located. For another pair, the expected values of the paths are different. In the test blocks, though, the starting and goal positions are fixed and the corresponding decision tree (in re-planning) is the same. 
    
The hallmark of our design was that the grid environment was not deterministic. Thus optimal planning was not \textit{necessarily} reduced to finding the shortest path but to select the path with the maximum expected value through experience. For instance, in Fig.~\ref{MM01}, top panel, if the starting point was at cell $3$ and the goal was at cell $27$, the shortest path, path A, had the lowest expected value compared to path B and path C. In the re-planning trials when path B was blocked, the second optimal path was path C which was longer than path A. 
  
\subsubsection{Results}
All three experiments that we designed had $6$ blocks of learning without payment (training phase). This allowed participants to explore the grid world without any concerns about their score. So if their scores became negative - as long as they were in the first $6$ blocks - they were not punished. At the beginning of these blocks, participants did more exploring and less planning but as they moved forward, they should be acquainted with the environment and able to find the optimal path. In order to examine this, we fixed the starting position at cell $3$ and the goal at cell $27$ for the last five trials of the $6$th block. With this design, we were able to recognize preliminary evidence of learning in the final blocks of the learning phase, specifically in block $6$. A participant who learned the grid world correctly, should be able to select path B (the optimal path) in these final trials otherwise we excluded his data. With this criterion, one of the participants' data was not included in our analysis. 

\begin{figure}
\includegraphics[scale=.45]{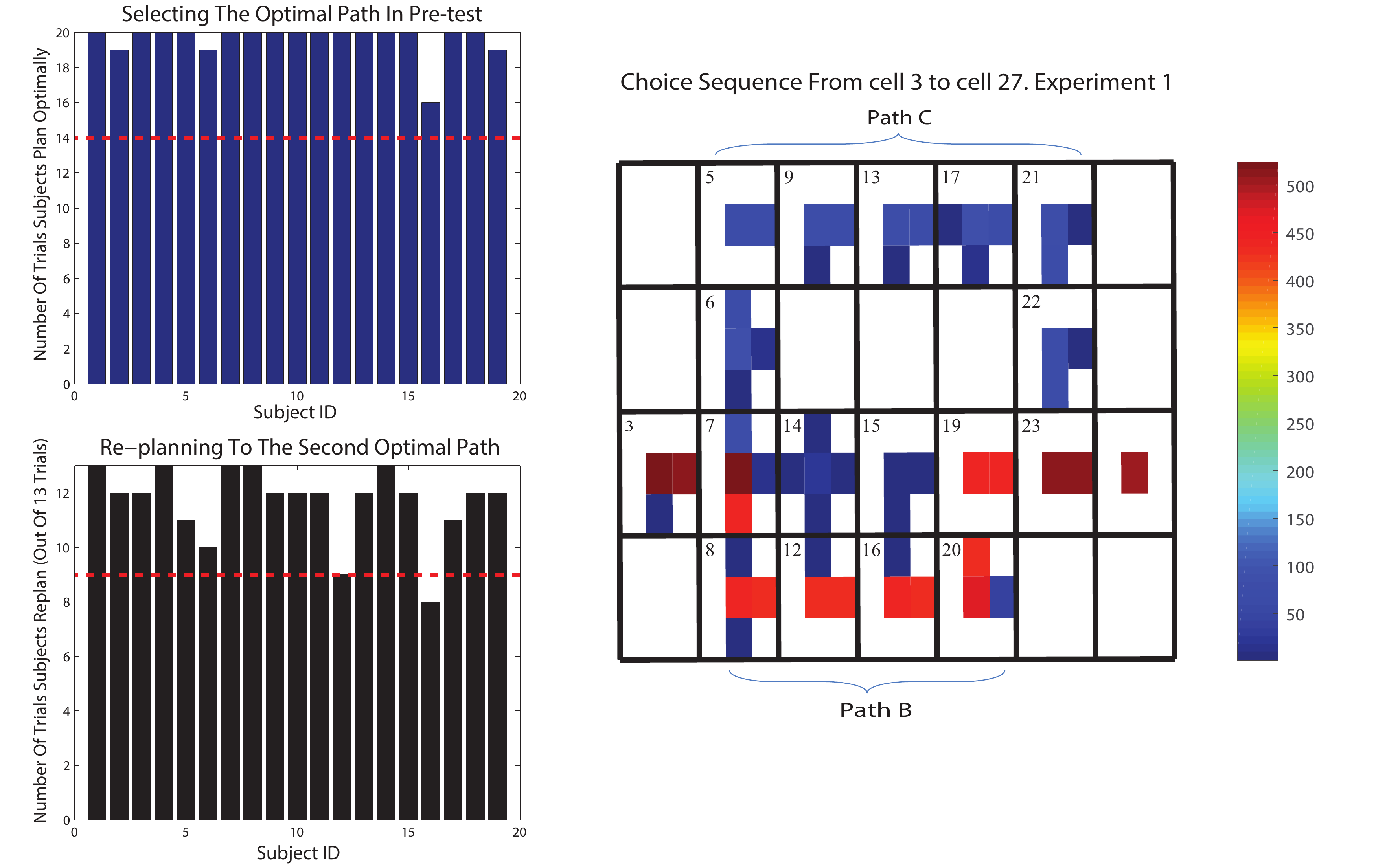}
    
\caption{{\bf Experiment $1$. Up left: All participants (total $19$), chose the optimal path ($\alpha = 0.05$) in the pretest block with $20$ different pairs of goal and starting points. Down left: $17$ participants re-planned to path C (the second optimal path) when they encountered with the blocked path B ($\alpha = 0.05$). The optimal path was blocked in $13$ trials (out of $40$ trials) in the test blocks. Right: Participants' choice in planning trials in the test blocks in experiment $1$. The heatmap shows the frequency of choices in the planning trials when path B was not blocked. Path B was the optimal path and selected frequently (depicted in red). The related cells were $8$, $12$, $16$, $20$, $19$ and $23$. The white cells with no cell number are obstacles.}} \label{Res03}
\end{figure}

Analyzing the pretest block's data (seventh block), all $19$ participants learned to choose the optimal path to reach the goal, Fig.~\ref{Res03}, up left. We used the binomial distribution to compute the minimum number of trials in which the participant must plan optimally to be able to reject the null hypothesis that the optimal path was selected randomly, ($\alpha=0.05$). Participants who chose the optimal path in only $14$ trials or less (out of $20$ trials) did not pass this test and failed to perform optimally. The red dashed line in top left panel of Fig.~\ref{Res03} shows this minimum.

In the test phase, path B was blocked in $13$ trials (out of $40$ trials). The minimum number of trials required for an optimal performance which could reject the null hypothesis was $10$. The red line showed this threshold. Participants who selected the (optimal) detour path less than or equal to $9$ trials out of $13$ trials were not optimal in re-planning trials. Only two of the participants failed to re-plan correctly to the second optimal path (path C), Fig.~\ref{Res03}, down left. Mixed-effect regression analysis with block number as regressor and participants as the random effect, showed that the average reward learning curve (as a function of block) was significant ($t(113) = 12.138$, $p<0.0001$).  Fig.~\ref{Res03}, right, shows that participants (on average) took path B, the optimal path, in planning from cell $3$ to cell $27$ more frequently than path A or path C. Note that path C was the second optimal path if path B was blocked. In the heatmap, the red color shows the more frequent directions or state occupancies. For each cell, the number of times participants visited a cell and the direction they took are counted. In the test blocks of experiment $1$, there are $27$ trials that re-planning did not occur and the starting and goal positions are fixed at $3$ and $27$. Ideally, if all participants choose the optimal path (path B) in all trials, cell $7$, $8$, $12$, $16$, $20$, $19$ and $23$ should be visited $19 \times 27 = 513$ times while path A and path C should not be selected. As the heatmap shows, there are few trials (less than $50$ trials in total) that path B is not selected (indicated by dark blue cells in path A and path C). 
  
\subsection{Experiment 2}
\subsubsection{Participants}
Thirty six healthy participants ($18$ males), performed experiment $2$ for payment of $\$9$ to $\$16$ based on their performance in the task. All participants signed the informed consent forms. Indiana University Institutional Review Boards has approved the study. 
    
\subsubsection{Task} \label{Exp2}
\begin{figure}
	\includegraphics[scale=.5]{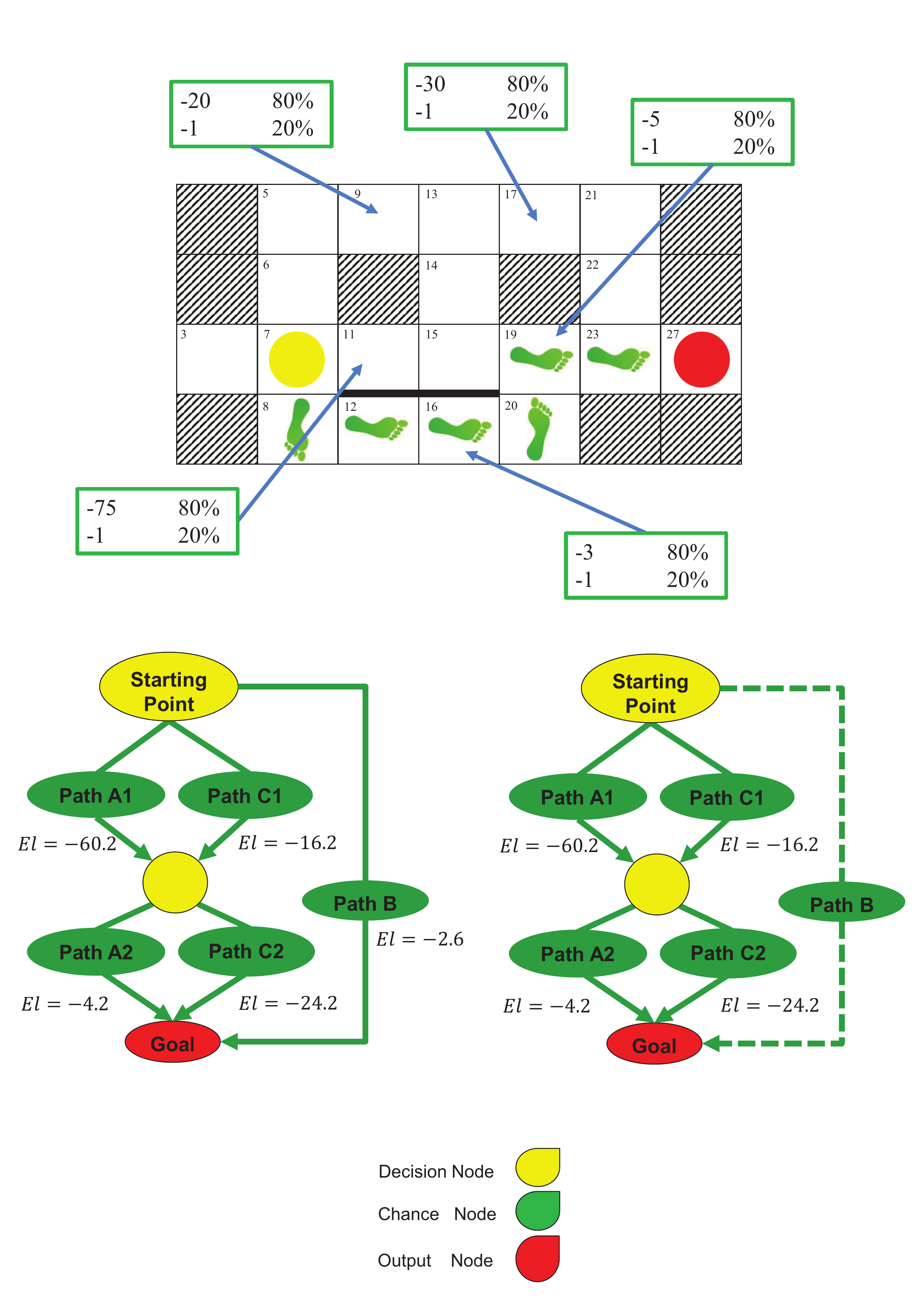}
	
	\caption{{\bf Experiment 2. Up: A 4 by 7 grid world experiment. The starting position is the yellow circle (cell $7$) and the goal is located at the red circle (cell $27$). Green footsteps indicate the optimal path from the starting point to the goal in the environment (which is path B). There are $5$ possible paths between the starting point and goal position. Path A1A2 through $7$, $11$, $15$, $19$, $23$, $27$; Path A1C2 through $7$, $11$, $15$, $14$, $13$, $17$, $21$, $22$, $23$, $27$; Path B through $7$, $8$, $12$, $16$, $20$, $19$, $23$, $27$; path C1C2 through $7$, $6$, $5$, $9$, $13$, $17$, $21$, $22$, $23$, $27$ and path C1A2 through $7$, $6$, $5$, $9$, $13$, $14$, $15$, $19$, $23$, $27$. Down: Corresponding decision tree for planning (left) and re-planning (right) with the exact $5$ possible paths between the start and goal positions. In the re-planning trials (on the right), path B is blocked in $30\%$ of the trials (dashed line). Expected losses are calculated based on the current starting point (cell $7$).}} \label{MM02}
\end{figure}

In the first experiment the best and worst paths (path B and path A) included a cell with a stochastic reward. In order to select the second optimal path in the re-planning trials, participants needed to compare a deterministic path (path C) with a stochastic path (path A). This might raise this question that our findings in the previous section only reflected participants' preference for a sure thing over a gamble. Therefore, we changed the environment in experiment $2$ and designed a more complicated environment as depicted in Fig.~\ref{MM02}. Similar to the previous experiment most cells had a regular punishment of $-1$, but there were $5$ cells with stochastic reward. These cells were $9$, $11$, $16$, $17$, $19$ and with probability of $0.8$ subjects received $-20$, $-75$, $-3$, $-30$ and $-5$ respectively. The number of available paths was increased from $3$ to $5$ and they were: (e.g. starting at cell $7$) Path A1A2 through $7$, $11$, $15$, $19$, $23$, $27$; Path A1C2 through $7$, $11$, $15$, $14$, $13$, $17$, $21$, $22$, $23$, $27$; Path B through $7$, $8$, $12$, $16$, $20$, $19$, $23$, $27$; path C1C2 through $7$, $6$, $5$, $9$, $13$, $17$, $21$, $22$, $23$, $27$ and path C1A2 through $7$, $6$, $5$, $9$, $13$, $14$, $15$, $19$, $23$, $27$. Path B which includes cell $16$ with expected loss of $-2.6$ is the optimal path. Choosing any path that contains cell $11$ with expected loss of $-60.2$ has the worst effect on participants' monetary reward (path A1A2 and A1C2). The losses at cell $9$, cell $17$ and cell $19$ were adjusted to create two sub-optimal paths: path C1C2 and C1A2. Based on the expected losses of these five critical cells, these paths could be prioritized as follows: path B, path C1C2, path C1A2, path A1A2 and path A1C2.

In this design the corresponding decision tree had two decision nodes and five chance nodes, the bottom panel of Fig.~\ref{MM02}. Similar to experiment $1$, in the test phase, the starting point and the goal position were fixed at $3$ and $27$ respectively. On $33\%$ of the trials in the test phase, when the optimal path is randomly blocked, path C1A2 was the optimal path among the remaining paths. Fig.~\ref{MM02}, b shows the corresponding decision tree.

\subsubsection{Results}
\begin{figure}
\includegraphics[scale=.6]{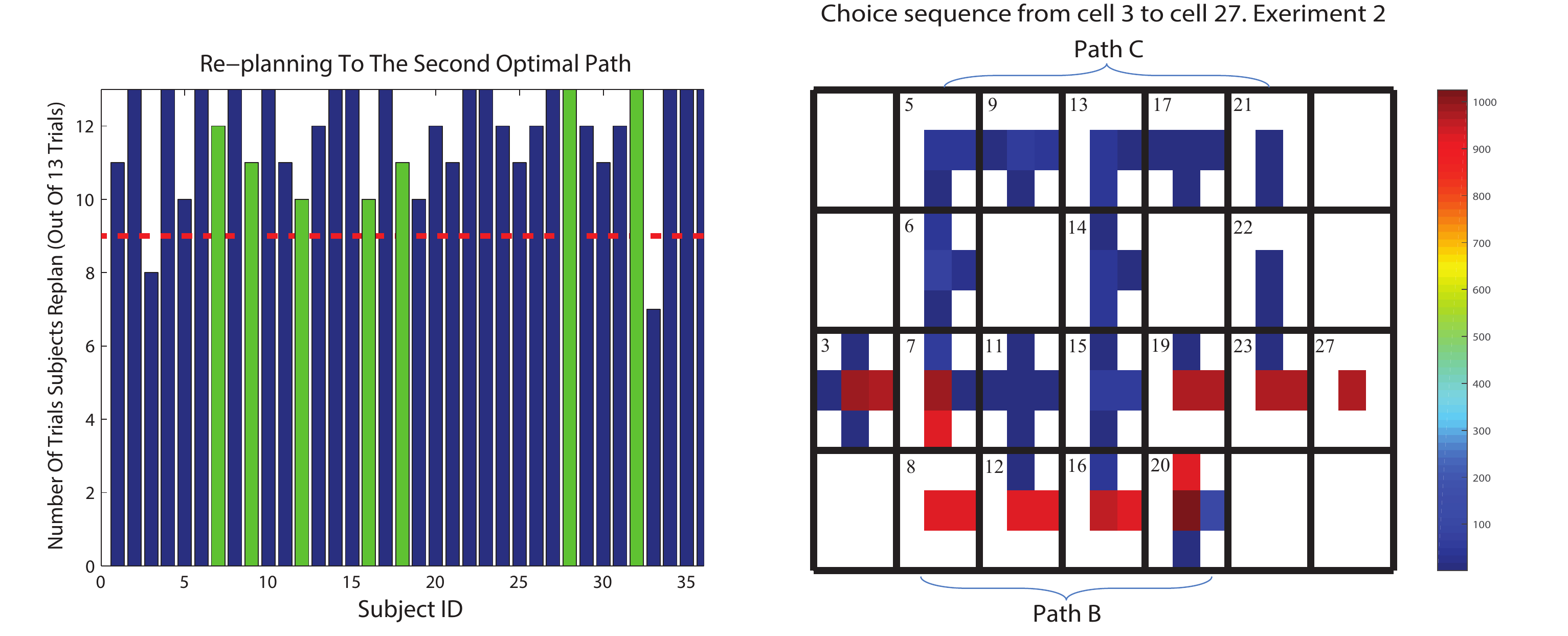}
  
\caption{{\bf Experiment $2$. Left: Blue bars show participants who correctly re-planned to the second optimal path which was path C1A2 (except the one below the red threshold) and green bars are the group of participants who re-planned to C1C2, the third optimal path, ($\alpha = 0.05$). The optimal path was blocked in $13$ trials (out of $40$ trials) in the test blocks. The goal and starting points were fixed but the detour path contained stochastic punishments. Right: (Heatmap) Participants' choice in planning trials in the test blocks in experiment $2$. The heatmap shows the number of times participants selected the optimal path in the planning trials when path B was not blocked. Path B went through the following cells: $7$, $8$, $12$, $16$, $20$, $19$, $23$, $27$. The blue cells shows path C1C2 and C1A2. A cell with darker red color is visited more often.}} \label{Res04}
\end{figure}

To analyze participants' learning curve, we fitted a mixed-effect regression to average reward that each participant earned during the first $7$ blocks (the learning and the pretest blocks). We found that the average reward learning curve as a function of block was significant ($t(215)=9.441$, $p<0.0001$). Note that the number of available paths increased but similar to experiment $1$, first, we checked whether the selected path was an optimal path or not. This is a simple straight forward test to check the optimal behavior in our experiment. As we did in experiment $1$, we computed the minimum number of trials in which the participant needed to choose the optimal path to reject the null hypothesis. Since the number of trials in the pretest block and the test blocks has not changed, these thresholds were similar to experiment $1$ which was $14$ in the pretest block and $10$ in the test blocks. Using this analysis, $30$ out of $36$ subjects found the optimal path in the pretest block ($\alpha = 0.05$). When it came to find the detour option, $27$ participants found the second optimal path correctly; $7$ chose the third optimal path and two did not re-plan at all, as shown in Fig.~\ref{Res04}, panel on the left. In our analysis, the optimal strategy was to first check the original optimal path (path B) and if it was blocked then switch to the second optimal path (path C1A2). Fig.~\ref{Res04}, right panel, captured this result from participants' choice. The heatmap shows the number of times participants visited each cell during trials when optimal path is not blocked in the test blocks ($27$ planning trials). Since we have $36$ participants, the maximum number of times that a cell could be occupied is $27 \times 36 = 972$ (represented in dark red). In few trials, non-optimal paths (path A1, path C1 and path C2) are selected (shaded in dark blue).
  
\subsection{Experiment 3}
In the instructions of experiment $1$ and $2$, participants were told that only one new obstacle would be added to the environment and it would block one path. Also in the test phase of the first and second experiments, the starting point was at $3$ and the goal point was at $27$ in all trials. But having a fixed pair in the test phase might indirectly lead participants to perform optimally in the re-planning trials. For instance, learning to go to cell $27$ from cell $3$ was easier than learning to go to cell $13$ from cell $5$ in pair (5, 13). In the latter pair, participants not only needed to know the stochastic payoffs but also paths' length to choose the optimal path. One should know that path $5$, $9$, $13$ had only $2$ steps with expected loss of $-17.2$ and path B had $10$ steps with two smaller losses (the total expected loss was $-13.8$). However in pair (3, 27), comparing the reward structures was enough to select the optimal path because path A was the shortest with minimum expected value and paths C1A2, C1C2, A1C2, A1A2 had smaller expected value and were longer than path B. In short, learning to choose the optimal path in some pairs was harder; consequently re-planning in these pairs when the environment changed (e.g. blockage in the optimal path) was more challenging.

Thus in experiment $3$, we changed the test phase ($8$th and $9$th block) to investigate this new question: how people modify their decisions when the change in the environment is less informative. Specifically, instead of having a fixed pair of starting point and goal position in the $8$th and $9$th blocks, we randomized these pairs (similar to what participants got used to do in the first $7$ blocks).

\subsubsection{Participants}
Among thirty two healthy subjects who participated in experiment $3$, there were $19$ males. Experiment $3$ has three blocks of test and thus the upper limit of payment is increased to $\$20$ but still depends on their performance in the task. All participants signed the informed consent forms. Indiana University Institutional Review Boards has approved the study. 
  
\subsubsection{Task} \label{Exp3}
In sum, we kept the same design as previous experiments but participants saw random pairs in the test phase. In order to keep the rate of changes in the environment (in the test phase) to $33\%$ (as it was in experiment $1$ and $2$), we increased the number of re-planning blocks to three blocks ($8$th, $9$th and $10$th) and therefore in $20$ out of $60$ trials the optimal path was blocked.    
  
\subsubsection{Results}
\begin{figure}
\includegraphics[scale=.45]{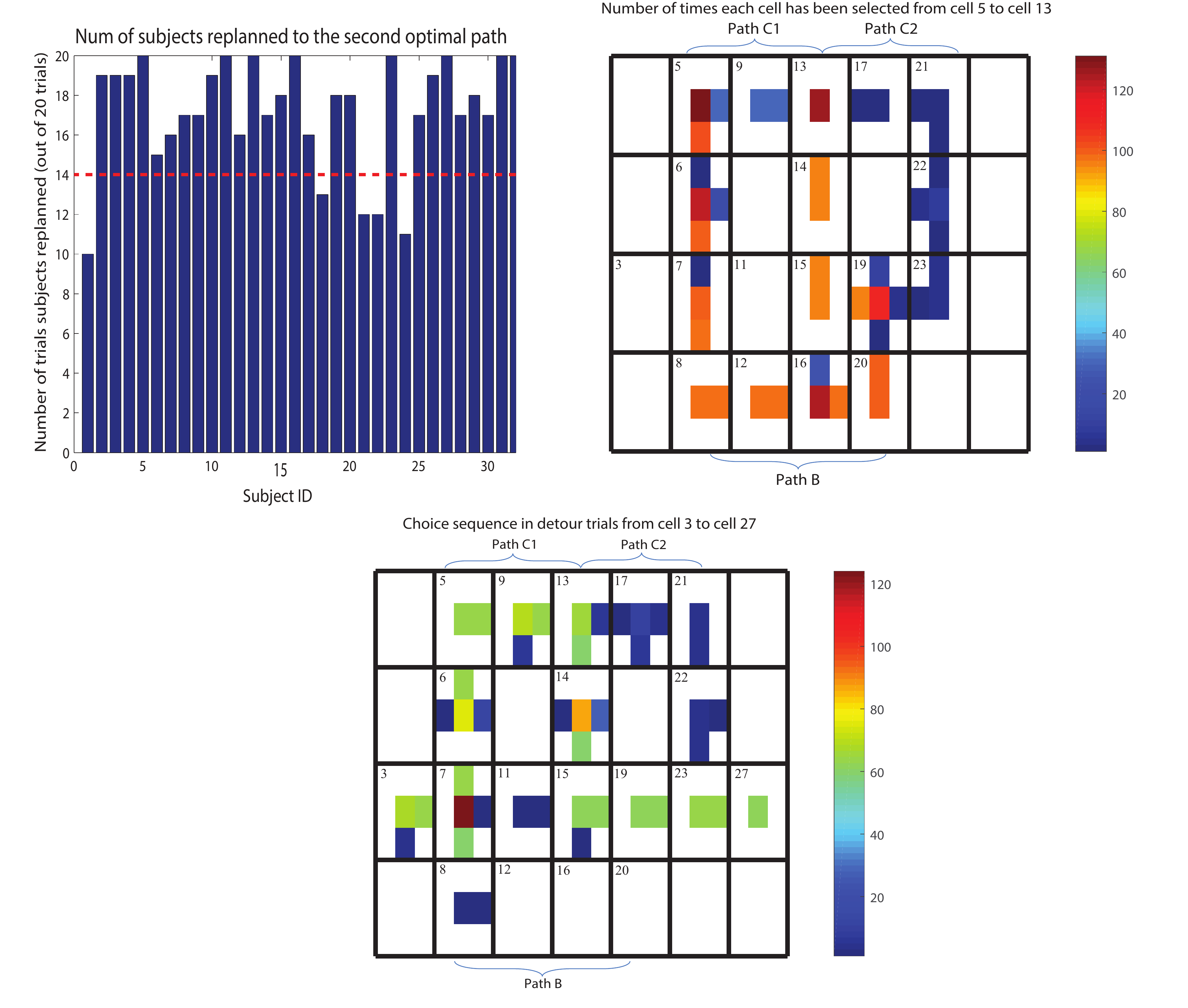}
  
\caption{{\bf Experiment $3$. Left: Blue bars above the red line indicate participants who could re-plan optimally. The optimal path was blocked in $20$ trials (out of $60$ trials) in the test blocks. Right: Heatmap of participants' choice in the planning trials in the test blocks in experiment $3$ when the starting position was at $5$ and the goal was at $13$. Pair (5, 13) was one of the challenging pairs in our design, which ruled out suggested models. The heatmap shows that majority of participants chose to take path B instead of going directly to cell $13$ using path C1. There was a (possible) decision node when they reach cell $19$. A few chose to take path C2 and were faced with the loss at cell $17$. Bottom: The number of times participants visited each cell in the grid when path B was blocked for pair (3, 27). Cell $12$, $16$ and $20$ are part of path B and that is why they never get selected in re-planning trials. The second optimal path went through $7$, $6$, $5$, $9$, $13$, $14$, $15$, $19$, $23$ and $27$ and was selected more frequently. Pair (3, 27) was repeated $6$ times. Note that white cells with no number are obstacles.}} \label{Res05}
\end{figure} 
  
Recall that in this experiment, in the test phase, the goal and starting point were not fixed, and thus for each pair we determined the optimal and non-optimal paths separately. We found that out of $32$ participants, there were  five who were able to find the optimal path in the pretest and the test trials, but not able to do so in the re-planning trials. Also there are two other participants that were able to plan optimally in the pretest and re-plan in the test phase but failed to choose the optimal paths in the test phase where there is no blockage. And the rest behaved optimally in both pretest and test blocks for planning and re-planning trials, see Fig.~\ref{Res05}, left panel.
   
We had $20$ different pairs in the pretest block (pair (7, 13) was different from pair (13, 7)). Only two participants did not plan optimally in this block. For some pairs, e.g. pair (13, 21), finding the optimal path was more challenging. In this particular pair, the non-optimal path was $13$, $17$, $21$ with stochastic punishment at cell $17$ with expected loss of $-24.2$ and the optimal path was the longer path of $13$, $14$, $15$, $19$, $23$, $22$, $21$ with expected loss of $-4.2$ at cell $19$. Participants needed to compare the longer path and greater expected value with the shorter path and smaller expected value. At pair (5, 13), the optimal path, path B, was even longer and less appealing. On average participants showed mediocre performance (below chance) in this pair. At pair (14, 22), though, they had no problem finding the optimal path probably due to the fact that path $14$, $13$, $17$, $21$, $22$ and path $14$, $15$, $19$, $23$, $22$ were symmetric with equal length and participants just needed to compare each path's expected reward. 
  
Fig.~\ref{Res05}, right panel, shows how often participants chose the optimal path in the test phase when it was not blocked (inspecting the planning behavior) averaged across all participants from cell $5$ to cell $13$. The heat map shows that participants chose to go down, heading to the optimal path, in spite of the fact that the shorter path requires only two steps. The panel at the bottom demonstrates re-planning behavior from cell $3$ to cell $27$ in the test phase with fewer data points. Note that in the test phase of experiment $3$, each pair was presented $6$ times and in only $2$ of them the optimal path was blocked. Thus the maximum number of times to choose the optimal path for all participants in experiment $3$ for each pair is $32 \times 6 \times \frac{2}{3} = 128$.

The bottom panel of Fig.~\ref{Res05} shows participants' choice in the re-planning trials for pair (3, 27) summarized as follows: a) Participants chose path C1A2 more than path C1C2 when path B was blocked. b) Cell $8$ should have been visited $2 \times 32 = 64$ times (light green) if all participants behaved optimally. c) Similarly, the total number of times that cell $7$ should have been visited is $128$ (going from $3$ to $7$ and then going from $8$ to $7$). In our data, it is $124$ due to the fact that there were trials that participants selected path C1 without checking path B and thus visited this cell once. d) Cell $6$ and cell $14$ have been visited more than $64$ times ($75$ and $87$ times respectively). We found that some participants failed to remember the correct transition matrix when they experienced a blockage in the optimal path and hit the surrounding obstacles two or three times before selecting the correct direction.

With the salient loss of $-75$ (with probability of $0.8$) occurring at cell $11$, we compared the rate that participants received this loss before and after the pretest block. The rate of experiencing $-75$ before the pretest block was $12.5$ while this rate went down to $1.25$ during the pretest and test blocks in experiment $3$  ($t(31) = 9.241$, $p<0.0001$). We should emphasize that in most of the trials before the pretest block, the path containing the $-75$ punishment was one of the accessible paths from the starting point to the goal position and more importantly in $60\%$ of these trials ($73$ out of $120$ trials) this path was the shortest path or one of the two equally shortest paths. The fact that the rate of selecting this path was reduced significantly from the learning blocks to the test phase showed that participants learned to avoid the great losses as they entered to the pretest and test blocks. We did the same analysis on the cells that deliver losses greater than $-5$. The rate of hitting cell $17$ with $-30$ loss in the learning blocks was significantly different from the test blocks ($t(31) = 6.253$, $p<0.0001$), but for cell $9$ with loss of $-20$, there was no significant changes. Note that all of these losses are probabilistic. From the mixed-effect regression analysis on the average reward learning curve (in the learning blocks), we found that the block number as a predictor was significant ($t(191) = 6.478$, $p<0.0001$). 

In the test blocks, participants were faced with random blockage of the optimal path, which might weaken their performance in recalling the obstacles \footnote{Besides the burden of recalling grid world's configuration, starting from the $7$th block, participants actually earned monetary rewards based on their choices and this might impair their performance and memory recollection.}. We used the mixed-effect regression to fit the number of times a participant hit an obstacle as a function of block number in the pretest and test blocks (considering participants as the random effect) and we did not find any significant difference between the rate of hitting obstacles in the pretest block vs. the test blocks ($t(2527)=1.426$, $p=0.154$) \footnote{Note that the obstacles were deterministic (in contrast with probabilistic punishments).}.\\

\section{Theoretical Analysis}
Reinforcement learning (RL) algorithms have been previously used to explain human behavior during learning of dynamic decision making tasks \cite{simon_neural_2011}, \cite{huys_bonsai_2012}, \cite{huys_interplay_2015}, \cite{sutton_reinforcement_1998} and can be formulated within a Markov decision process (MDP) framework. In a MDP, the communication between an agent and the stochastic environment (at time step, $k$) is through three sets defined by state $\left \{s^k \right \}$, action $\left \{a^k \right \}$ and (probabilistic) reward $r_k$. After taking action $a_k$ the environment transfers to a new state $s_{k+1}$ with probability $\textbf{T}_{ij}^u (k)=\Pr{(s_{k+1}=s^j \lvert s_k=s^i,a_k=a^u) }$ and the agent receives a probabilistic reward $r_k=r$ with probability $\textbf{R}_{ij}^u (r,k)=\Pr{(r_k=r \lvert s_k=s^i,s_{k+1}=s^j,a_k=a^u) }$, \cite{khodadadi_learning_2014}. The transition function and the reward function have the Markov property meaning that the transition probability and the reward probability are independent of the history of states and actions $\left\lbrace s_1,a_1,\cdots,s_k,a_k\right\rbrace$ and merely depend on the state and the action at time step $k$. In other words:
   
\begin{equation}
\Pr{(s_{k+1} = s^j \lvert s_k,a_k,\cdots ,s_1,a_1)}=\Pr{(s_{k+1}=s^j \lvert s_k,a_k) } \label{eq:08}
\end{equation}
   
\begin{equation}
\Pr{(r_k=r \lvert s_{k+1},s_k,a_k,\cdots ,s_1,a_1)}=\Pr{(r_k=r \lvert s_{k+1},s_k,a_k) } \label{eq:09}
\end{equation}

The main goal is to find the optimal policy, $\pi^*$, which determines the probability of selecting action $a$ in state $s$ while maximizing a desired function of accumulated rewards called \textit{return}. A widely used form of the return function is the \textit{expected sum of future reward discounted by $\gamma$}, \cite{sutton_reinforcement_1998}: 
  
\begin{equation}
\mathrm{E} \left[ \sum_{k=0}^{\infty} \gamma^k r_k \right] \label{eq:01}
\end{equation}
  
where discounting factor, $\gamma$, weighs the future reward relative to the immediate reward. Using the return function in equation~\ref{eq:01}, \textit{state-action} function, $Q(s, a)$, can be defined as the expected discounted sum of rewards received in state $i$ and action $j$ at time step $k$ given the policy, $\pi$:
  
\begin{equation}
Q_{\pi} \left( s^i, a^j, k\right)=\mathrm{E} \left[ \sum_{u=k}^{\infty} \gamma^{u-k} r_u \lvert s_k=s^i, a_k=a^j, \pi \right]
\label{eq:02}
\end{equation}
 
\begin{equation}
Q_\pi(s_k, a_k) = \mathrm{E} \left[ r_k+\gamma Q_\pi (s_{k+1}, a_{k+1}) \right] \label{eq:03}
\end{equation}
  
This equation is known as the Bellman equation. The expectation on the right hand side of this equation depends on the functions $\textbf{T}$ and $\textbf{R}$. Notice that the Bellman equation provides one equation for each state and so can be considered as a set of recursive equations or more formally a system of equations. If the transition and reward functions are known, the Bellman equation can be solved by dynamic programming method, \cite{bertsekas_neuro-dynamic_1996}. However, in most cases, including our experiments, the agent does not know the model of the system. 
  
The model-based method tries to estimate $\textbf{T}$ and $\textbf{R}$ functions and solve the Bellman equation with \textit{value iteration}, \cite{sutton_reinforcement_1998}, \cite{daw_uncertainty-based_2005}. In this approach, the model-based agent learns the model of the system and uses this model to find the optimal policy. However, the model-free approach such as temporal difference learning algorithm, uses an estimate of the difference between the two sides of Bellman equation~\ref{eq:03} which is called the temporal difference (TD) error. Therefore, \textit{without} knowing the $\textbf{T}$ and $\textbf{R}$ functions, TD is able to learn the optimal policy which can maximize the return function. It is also computationally simple and efficient. However, its ascendancy is limited to a stationary environment. Since it \textit{locally} updates the state-action function (equation~\ref{eq:03}), if there is a change in the environment (e.g. re-planning in our experiments or reward devaluation) it takes a long time to propagate that (new) information to update all the state-action estimations. In addition, in the training blocks of our experiments, the TD algorithm can not learn the correct state-action function or learn the optimal path because the goal and starting point are not fixed. At best it can learn different sets of state-action function for each pair and for each pair retrieve that relevant information (which is reduced to a look up table). Because of these reasons, the model-free RL (model $1$) is not a good candidate to analyze our results, but to complete our model comparison benchmark, we documented the quantitative fit. 

The simplest form of the temporal difference (TD) error is defined in equation~\ref{eq:04} and is used to update the state-action function which is called the Q-learning model (equation~\ref{eq:05Q}):
  
\begin{equation}
\delta_k = r_k + \gamma \cdot max_a \hat{Q}_{\pi,k}(s^{u}, a) - \hat{Q}_{\pi, k} (s^i, a^j) \label{eq:04}
\end{equation}

\begin{equation}
\hat{Q}_{\pi, {k+1}} (s^i, a^i) = \hat{Q}_{\pi,k} (s^i, a^i) + \alpha_c \cdot \delta_k \label{eq:05Q}
\end{equation}
  
Note that $\alpha_c$ is the learning rate. Because the goal and starting points could be in any side of the grid, the associated $Q(s,a)$ is initially set to $Q_0$ for all cells in the grid, \cite{simon_neural_2011}. We could have specified different $Q_0$ for each cell and for each action according to their position in the grid and the starting position (and especially if they were next to the grid's borders), but this would add more free parameters to the model with negligible benefit. Using the Q-values, an action selection module, e.g. Softmax decision rule, can be used to choose the appropriate action with a probability of:
  
\begin{equation}
p_k(a) = exp(\beta\cdot Q_k(s,a))/\sum exp((\beta\: \cdot Q_k(s,a)))
\label{eq:08tem}
\end{equation}
  
where $\beta$ is the temperature. If $\beta \rightarrow 0$, the algorithm selects a random action (in our case, we have $4$ actions: up, down, right and left, thus each action is selected with probability of $0.25$) and if $\beta \rightarrow \infty$, the action with the maximum Q-value is selected (as in $\epsilon$ -greedy policy). 

We also used a {\bf baseline model}, with no learning. A baseline model (model $1$) randomly chooses an action at each state and does not consider the previous experiences. Unlike the two models that we explained above, modeling participants' behavior in our task is not trivial. A candidate model should be able to plan ahead and estimate the reward function simultaneously and appropriately. With these goals in mind, the baseline model has no planning at all and fails to learn the payoffs in the grid world and the model-free RL has problems in learning the correct reward function and planning because of the dynamically changing environment. Again, the model-free RL plans for one step-reward and with changing environment this strategy is not useful. In the following, we explain how different strategies of estimating the reward function and various manipulations of the planning module can change models' predictions in the learning and test phases.
  
\begin{figure}
\centering
\includegraphics[scale=.5]{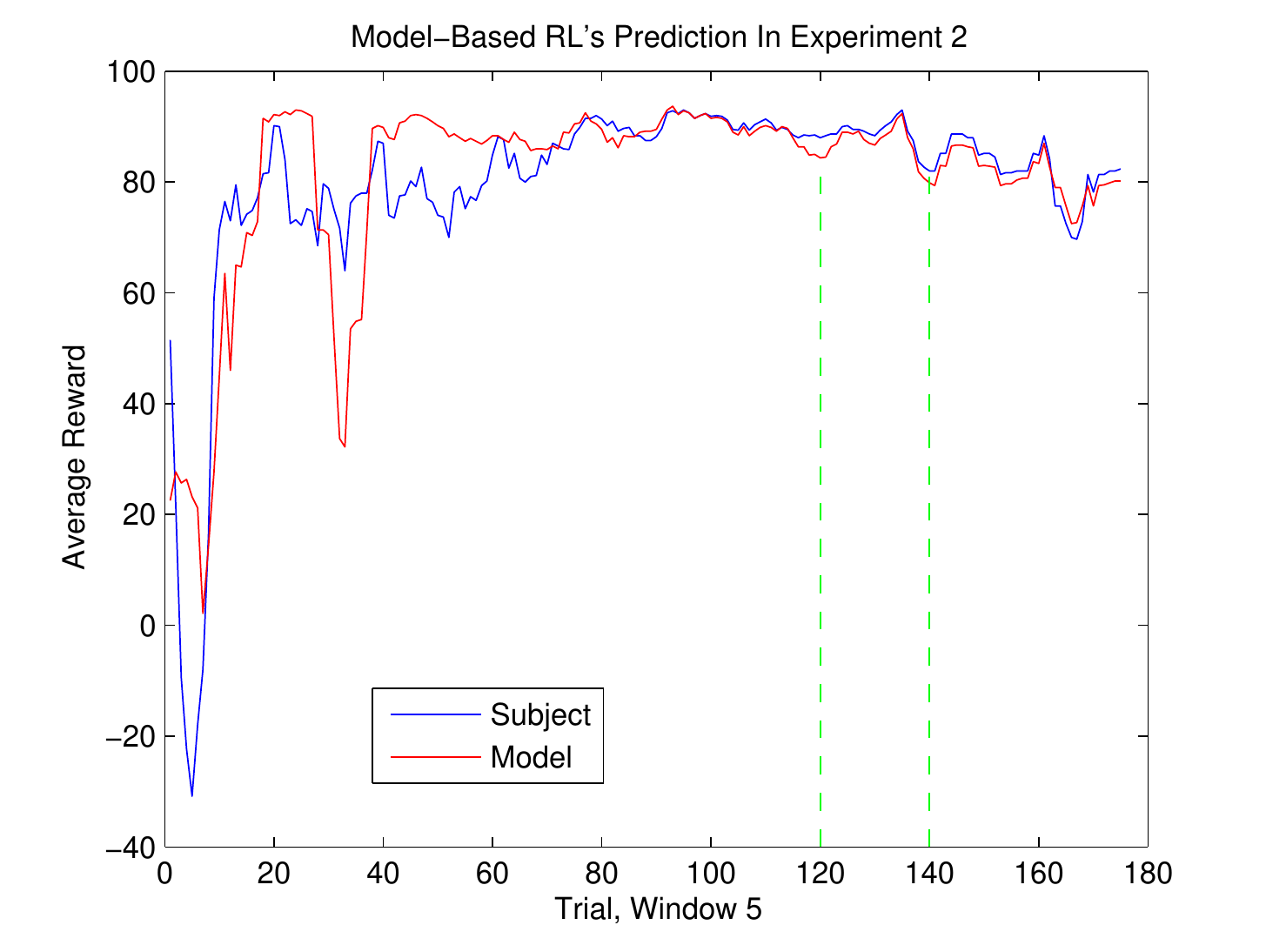}
\caption{{\bf Model-based RL's prediction in (red) for experiment $2$. Green lines indicate the beginning of the pre-test block ($7$th block) and the test blocks ($8$th and $9$th). Data related to subject $14$.}} \label{Res07}
\end{figure} 
  
\subsection{Model Based Reinforcement Learning}  
As discussed earlier, the model-based RL solves the Bellman equation by learning the model or representation of the environment (cognitive map in \cite{tolman_cognitive_1948}). Unlike TD learning algorithm, the model-based RL's estimates and updates are \textit{global} and because of this distinction, when the environment changes or the reward is devalued, the model-based RL can respond to these changes quickly. We compute the state-action function with q-value iteration (\cite{sutton_reinforcement_1998}, \cite{bertsekas_neuro-dynamic_1996}) as follows: 
  
\begin{equation}
\hat{Q}_{k+1}(s^{i}, a^{u})\leftarrow \sum_{s^{j}} T(s^{i},a,s^{j})[R(s^{i},a,s^{j})+ \gamma max_a (\hat{Q}_{k}(s^{j}, a))] \label{eq:07}
\end{equation}

Generally estimating $\textbf{T}$ and $\textbf{R}$ function depends on the environment with which the agent is dealing. Since the obstacles in our grid world are deterministic, with only one experience, participants can learn the transition function. We can also assume that each time an action is selected in a specific state, the transition function is updated by a learning rate respectively. In order to learn the probabilistic reward function, one possible method is to take the average of what participant has earned for each state-action pair so far (will be discussed later in equation~\ref{eq:10} and ~\ref{eq:11}). Using these two pieces of information, we can compute the state-action value function ($Q(s,a)$)  at \textit{each} state with q-value iteration (\cite{sutton_reinforcement_1998}, \cite{bertsekas_neuro-dynamic_1996}). Once we have state-action function, we choose the appropriate action using Softmax decision rule as in equation ~\ref{eq:08tem}.

The average reward predicted by model-based RL for experiment $2$ is shown in Fig.~\ref{Res07}. The model-based RL predicted similar behavior as our subject (no. $14$). To show the qualitative fit, we chose a very basic version of the model-based RL that updates its estimation \textit{at each step} using q-value iteration method. We assumed that the model knows the correct transition matrix and does not need to learn the $\textbf{T}$ function. 

Taking the average of previous samples requires a perfect memory (specifically at cells with stochastic rewards) and it is not practical for human participants. Instead, we use a linear filter to approximate the reward function dynamically using equation~\ref{eq:10}:
  
\begin{equation}
R(k+1) = \alpha_1 \cdot R(k) + (1-\alpha_1) \cdot r
\label{eq:10}
\end{equation}
  
where $r$ is the reward that is delivered upon transition and $R(k)$ is the averaged reward up to time step, $k$. $\alpha_1$ determines the relative importance of the averaged (observed) reward versus the current reward. We also include a forgetting factor, $\alpha_2$, which plays a decaying role in reward estimation (equation~\ref{eq:11}) when the greater loss is not experienced (e.g. in $20\%$ of the time, these cells have a regular loss of $-1$):
  
\begin{equation}
R(k+1) = \alpha_2 \cdot R(k)
\label{eq:11}
\end{equation}
   
Assuming that people are only sensitive to great losses, learning and estimating the reward function only occur when a great loss is experienced. Whenever a participant experiences a great loss (anything but $-1$) at a cell, the model updates its estimation using equation~\ref{eq:10}. We tested models with different assumptions regarding $\alpha_1$, $\alpha_2$ by categorizing the losses into three levels of saliency (low, medium and high): Cell $16$ and cell $19$ with possible losses of $-3$, $ -5$ (low saliency, LS), cell $9$ and cell $17$ with probable losses of $-20$, $ -30$ (medium saliency, MS) and cell $11$ with loss of $-75$ in $80\%$ of the time (high saliency, HS). Similar pattern can be formed for experiment $1$ with $-45$ at cell $21$ with medium salient loss, very salient loss of $-75$ at cell $15$ and a small loss of $-3$ at cell $16$. Based on Bayesian Information Criterion (BIC), we found that the model with $\alpha_{1_{HS}}, \alpha_{1_{MS}}, \alpha_{1_{LS}}$ and no decay rate was the best to use.\\

In addition to linear filter, there are other methods to estimate reward function as well (e.g. simple strategies or rules). In the next section, we introduce $3$ different heuristics that we use to learn the reward structure. Note that we are still using the model-based RL framework but each strategy estimates the $\textbf{R}$ function differently with small or no computational cost (model $4$, $5$ and $6$). 

\subsection{Heuristic-Based Models}  \label{Heuristic}
We use $3$ different heuristic-based models to fit to our data (training dataset). These heuristics are: avoiding the great loss, remembering the last reward and finding the shortest path. These heuristics along with previously described models are summarized in Table~\ref{tab:models}. \\

  \begin{table}[h]
  	\centering
  	\caption{Summary of models and their free parameters}
  	\label{tab:models}
  	
  	\begin{center}
  		\begin{threeparttable}
  			\begin{tabular}{clcc}
  				\toprule		
  				\textbf{Model} & \textbf{Description} & \textbf{Parameters} & \parbox[t]{2cm}{\textbf{No. of free\\parameters}} \\ \midrule
  				1 & Baseline (random model)  & - & 0\\[1pt]
  				2 & Q-learning  & $ \beta, \gamma, q_{u_0}, q_{r_0}, q_{d_0}, q_{l_0} $ & 6\\[1pt]
  				3 & Model-based RL  & $ \beta, \gamma, \alpha_{1_{HS}}, \alpha_{1_{MS}}, \alpha_{1_{LS}} $ & 5\\[1pt]
  				4 & Avoids Salient Loss -MBRL & $ \beta, \gamma $ & 2 \\ [1pt]
  				5 & Remembers The Last R -MBRL & $ \beta, \gamma $ & 2\\[1pt]
  				6 & Finds The Shortest Path -MBRL & $ \beta, \gamma $ & 2\\[1pt]
  				7 & Cubed Model-Based RL &  $ \beta, \gamma, \alpha_{1_{HS}}, \alpha_{1_{MS}}, \alpha_{1_{LS}}, \omega $ & 6 \\ [1pt]
  				8 & Successor Representation &  $ \beta, \gamma,  \alpha_l, \alpha_{1_{HS}}, \alpha_{1_{MS}}, \alpha_{1_{LS}} $ & 6 \\ [1pt]
  				9 & Avoids Salient Loss -SR &  $ \beta, \gamma, \lambda $ & 3 \\ [1pt]
  				10 & Remembers The Last R -SR &  $ \beta, \gamma, \lambda $ & 3 \\ [1pt]
  				11 & Finds The Shortest Path -SR &  $ \beta, \gamma, \lambda $ & 3 \\ [1pt]
  				12 & Hybrid SR-MB &  $ \beta, \gamma, \alpha_l, \omega_{hb}, \alpha_{1_{HS}}, \alpha_{1_{MS}}, \alpha_{1_{LS}} $ & 7 \\ \bottomrule
  									
  			\end{tabular}
  			  	\begin{tablenotes}
  			  		\footnotesize MB = Model-based, SR = Successor Representation.
  			  	\end{tablenotes}  			
  		\end{threeparttable}
  	\end{center}
  \end{table}

{\bf Avoiding the salient loss}: \\
As we discussed above, participants were sensitive to the punishment at cell $11$ with expected value of $-60.2$. This cell had the greatest and the most \textit{salient} loss. A candidate model which {\bf avoids the salient loss} can explain the behavior of our participants who chose not to enter to cell $11$ (model $4$). This model does not estimate the reward function but only avoids cell $11$ in experiment $2$ and $3$ (and cell $15$ in experiment $1$) with $-75$ loss. Note that this behavior does {\bf not} indicate that these participants chose the optimal path necessarily. For instance, in pair (5, 13), we had two groups of participants: The ones who failed to find the optimal path by choosing path C1 (and facing the loss of $-20$ at cell $9$) and not path B (with the minimum loss) and participants who learned about other punishments which are not as salient as cell $11$ and at the end chose path B (which is optimal). \\

{\bf Remembering the Last Reward}:\\  
The fifth model (in Table~\ref{tab:models}) is categorized as a model-based RL, but only remembers the last reward and does not estimate the R function. This model saves the {\bf last} value of the reward at each cell. Thus with one bad experience (e.g. receiving $-75$ at cell $11$), it is less probable for the model to enter that cell. The last value of any experience is saved for the rest of the computations and action selection. This model does not need the full memory of reward samples because it does not need to estimate or learn the stochastic reward structure. This simplicity can cause sub-optimal behavior in re-planning trials during the test phase. Consider the following scenario: In the 8th block at pair (7, 13), the optimal path (path B) is blocked, the model chooses path C1, assuming that the last experience of path A1 was with the great loss at cell $11$. Then at cell $9$, it would receive $-1$ instead of $-20$. This would be saved in the model's memory. For the next pair, e.g. pair (5, 13), which includes cell $9$, the model selects path C1 over path B simply because its last experience at cell $9$ is with a regular punishment. Moreover, it only takes two steps to catch the goal with the expected value of $99$. Path B, however, requires participants to take $10$ steps and its expected value is $88.4$. This strategy would be very useful when the environment is deterministic. \\

{\bf Finding The Shortest Path}:\\  
There are few participants that chose the shortest path (A1A2) in the trials with a blocked optimal path, but were able to find the optimal path in regular trials. This suggests that these participants did not care about maximizing their reward when they encountered an obstacle in the optimal path (against the goal of the experiment), and it is only important for them to catch the goal using the remaining available moves. This approach (model $6$), with no re-planning, {\bf finds the shortest path} instead of choosing a path with minimum loss. Also note that, choosing the shortest path, is not necessarily distinguishable from other strategies since there are so many pairs in these experiments that have two paths of equal length. For instance, in experiment $1$, in the pretest block, $12$ out of $20$ pairs have this characteristic: a path with a regular punishment and one with a great loss but both have equal length. Therefore, model $6$ can choose any of these paths with equal probability. 

In Experiment $1$, there were two people who chose the shortest path in those trials. In experiment $2$, there were $6$ participants who failed to choose the second optimal path and were not sensitive to loss $-30$ and one person who chose the shortest path ignoring all the losses. Finally in the third experiment, there were $7$ participants with similar pattern in their behavior. What surprised us in these results was that these participants learned to find the optimal path when they were asked in the pretest block but with experiencing the blockage in the optimal path, they ignored or forgot their findings on reward structure and switched their strategy from choosing the optimal path to the shortest path. \\

\subsection{Cubed Model-based RL: Planning On A Smaller Grid} 
Earlier, we explained how the shortest path and the optimal path are different in our design. Based on this distinction, we define the \textit{length} of each path as the number of steps in the shortest optimal path. Thus the length of the path between cell $5$ and cell $13$ is $10$ because the optimal path contains path B but it only takes two steps to reach cell $13$ from cell $9$ (disregarding the stochastic punishments). Now assume that at the beginning of the learning phase a participant is at cell $5$ and she wants to reach cell $13$. At the first glance, path C1 is the shortest and path B is the longest. But if she only takes one step, she would probably ($80\%$ of the time) faced with the great loss at cell $9$. Next time she would divert her decision to either path A1 or path B. Assuming she chose path A1, after three steps, she would experience the maximum loss at cell $11$ with probability of $0.8$. These two results would lead her to take path B in the next trial. In this scenario one could use q-value iteration for three steps ahead (instead of the whole path as in model 3) and prune the rest of the paths and lower the computational cost (model $7$). This model, which we called "cubed model-based RL", is fundamentally different from the previous heuristic-based models because the planning module has been changed. In fact, Model $7$ plans for a small window ahead of the current position (using the shortest length for each pair). 

The minimum number of steps to prune the decision tree can be an extra free parameter in cubed model-based RL. While pruning at three steps is good for pair (5, 13), it is not a good stopping point for pair (3, 17). In order to compare path C1 and B correctly, one should plan for at least four steps ahead from cell $3$ to be able to see the losses at cell $12$ and cell $9$. So for different pairs, the model needs to use different pruning depths. The diversity of pairs in each experiment (e.g. experiment $3$ has $65$ pairs) forces this model to pick the maximum number of available steps for planning (which is $7$ in our grid world) and thus with one extra parameter, cubed model-based RL can not be distinguished from model $3$ for many participants at many pairs as we report in the model fit result. 
  
So far we have discussed about the first $7$ models in Table ~\ref{tab:models}. These models are different in their planning and learning approaches. In traditional model-free RL, the Q-values are computed using TD errors. But since the goal is not fixed and the starting points are randomized, this approach can not learn the correct Q-values at each states. Moreover, TD mechanism is unable to plan ahead when the environment is not deterministic. The baseline model lacks learning and planning regardless of the environment. We proposed $3$ different heuristics (avoiding salient losses, finding the shortest path and remembering the last punishment) based on our data. Although these strategies are simpler to simulate and require less computations (because they do not estimate the {\bf R} function), they can \textit{only} explain a small portion of our data and mainly have wrong predictions. The (full) model-based RL, model $3$, learns the model of the environment and estimates the reward structure. Having the model of the environment, which is independent of the goal or the starting position, helps the agent to predict the consequence of each action before taking them. 

\subsection{Successor Representation}  
In addition to the main RL families that have been explained so far, there is another algorithm called successor representation (SR) that was first introduced by \cite{dayan_improving_1993} which has the flexibility of the model-based RL and the simplicity of the TD learning. In this algorithm, the Q-values are decomposed into a reward matrix and a successor map which predicts the (discounted) future occupancy of all states, $M$ (which is called SR matrix). Given any initial state, $s^i$, the SR matrix counts the number of times that a subsequent state (in a trajectory) is visited later:

 \begin{equation}
M(s^i, s^u, a^j) = \mathbb{E}\left [ \sum_{\nu=0}^{\infty }\gamma^{\nu} \mathbb{I}\left [s_{k+\nu}= s^u\right ] \Bigm| s_k=s^i, a_k=a^j \right ]
 \label{eq:M}
 \end{equation}

where $\mathbb{I}\left [ . \right ] = 1$ when its argument is true and zero otherwise, \cite{kulkarni_deep_2016}. While it is possible to compute the SR matrix from the transition matrix \cite{gershman_successor_2012}, it is more common (and less expensive) to estimate SR matrix via Bellman equation as we did in Q-learning and the updating rule is as follows ($\alpha_{l}$ is a learning rate):

 \begin{equation}
M(s^i, s^u, a^j) \leftarrow M(s^i, s^u, a^j) + \alpha_{l} \left [ \mathbb{I}\left [s_{k+1}= s^u\right ] + \gamma M(s_{k+1}, s^u, a_{k+1}) - M(s^i, s^u, a^j) \right ]
 \label{eq:BellM}
 \end{equation}

In order to calculate the Q-values given a policy $\pi$, we need to compute the inner product of the reward and SR matrix using equation~\ref{eq:qMSR}, \cite{kulkarni_deep_2016}:

 \begin{equation}
Q_\pi(s,a) = \sum_{s^u} M(s^i, s^u, a^j) R(s^u)
 \label{eq:qMSR}
 \end{equation}

As we discussed before, there are multiple methods to estimate reward function (three different heuristics and one linear filter, equation~\ref{eq:10}). Consistent with our model-based RL fit, we use the linear filter and heuristics to learn the reward function (avoiding the great loss, remembering the last R and finding the shortest path related to models $9$, $10$ and $11$ respectively). 

\subsection{Hybrid SR-MB}   \label{HybExplain}
It is important to emphasize that the SR matrix, $M$, is different from the transition matrix. Thus, if there are abrupt changes in the environment, i.e. transition revaluation, the SR model fails to adapt to these changes without learning the new trajectories. In our experiments, in the test blocks, the random blockage in the optimal path is one example of transition revaluation. In order to choose the correct action, the SR algorithm needs to relearn the (new) trajectories. One possible solution to this problem is to combine the SR model with the model-based RL model (hybrid SR-MB), \cite{momennejad_successor_2016}, \cite{russek_predictive_2017}. For instance, the probability of each action can be a weighted average of the model-based RL's predicted probability and the SR's predicted probability. By assigning a greater weight to the model-based RL's prediction, the hybrid SR-MB can be more flexible toward the sudden changes of the transition matrix.

\section{Model Fit Results}
Although we have $12$ different models, some of them share common characteristics in planning or reward estimation. For instance, model $5$ is different from model $4$ in estimating the reward function (model $5$ only remembers the last reward while model $4$ avoids the greatest loss), but they both use model-based decision making in planning. Because of these differences and similarities, we divide these models into $6$ unique categories: a) the random choice model with no planning and no learning (model $1$). b) the model-free RL which does not learn the structure of the environment but does plan based on one-step rewards and the next step's expected future reward (model $2$). c) the model-based RL and/with heuristics (including models $3$, $4$, $5$ and $6$). These models are similar in their planning but differ on how they estimate the reward structure. d) the cubed model-based RL (model $7$) with a constrained view of the environment. e) the SR model and/with heuristics (including models $8$, $9$, $10$ and $11$). These models learn a similar multi-step representation of the environment, but their estimations of the reward function are different. Note that the SR models are not as good as model-based RL in learning the environment but better than the model-free RL. f) the hybrid SR-MB (model $12$) which blends the SR algorithm with the model-based RL to improve its performance in transition devaluation.  

In some of these categories, i.e. category c, there are four different models and in some, i.e. category f, there is only one model. Our strategy to compare these models is to first nominate one candidate in each category (within group comparison) and then measure how good they are in contrast to other categories. Note that we first compare the performances of these models in the learning blocks and then evaluate their predictions in the test blocks.

\subsection{Model Fit Results In The Learning (Training) Blocks} \label{ModelFit}
There are multiple ways to evaluate models. The first that we chose to report is based on BIC comparison \cite{schwarz_estimating_1978}. To this end we fitted these models to the data from learning blocks (block $1$ to $6$ with $n=120$ trials) using one-step-ahead prediction method, equation~\ref{eq:LL}. In this method, the model predicts (each) participant's choice on the next trial using the sequence of choices and payoffs that she has taken and experienced respectively. Specifically, the candidate model calculates the probability of selecting a particular action, that the participant selected, at each cell using the history of choices and rewards/punishments ($\left\lbrace s_1,a_1,r_1\cdots,s_k,a_k,r_k\right\rbrace$). 
  
\begin{equation}
LL_{i}=\sum_{1}^{n-1}ln(Prob(a_{k+1}|\left\lbrace s_1,a_1,r_1\cdots,s_k,a_k,r_k\right\rbrace)) \label{eq:LL}
\end{equation}
  
In order to find the best fitted parameters, the maximum likelihood estimation is applied (simplex search algorithm) and BIC is computed as in equation~\ref{eq:BIC}:
  
\begin{equation}
BIC = -2\cdot ln(LL_i) + k\cdot ln(N) \label{eq:BIC}
\end{equation}
  
where $N$ is the number of observation and $LL_i$ is the likelihood of the $i^{th}$ participant. 

First, we compared the models in the model-based RL category (model-based RL and heuristics, category \textbf{c}). Model $5$ which remembers the last reward wins in experiment $1$, but model $3$, model-based RL with a linear estimation of reward, has the lowest BIC in experiment $2$ and $3$. Simplicity of the environment in experiment $1$ plays a critical role in this result. In experiment $1$ there were only three paths (rather than $5$) and the detour path (path C) had no stochastic punishment. The loss at path B, the optimal path, was very small and even if the participant was unlucky, the worse case loss was $-3$. In path A, however, the difference between the regular punishment and the worse case loss was considerable. In $80\%$ of the times, the environment delivered a loss of $-75$ and thus with one such experience the model with the strategy of remembering the last punishment was no longer likely to choose this path. Therefore, model $5$ was able to find the optimal path with only two free parameters (temperature and discount) without learning the reward structure. In the other two experiments because there was $20\%$ chance that participants receive a regular punishment of $-1$ at cell $9$ and cell $17$, the strategy of recording \textit{only} the last punishment might mislead the model to choose the wrong path. 

Similarly, the (within category) model comparison for SR models, the full SR model had the better BIC in experiments $2$ and $3$ while the heuristic-based SR model which ignores the rewards (and chooses the shortest path) won in experiment $1$ \footnote{Important to mention that in experiments $2$ and $3$, we had similar AIC results but not in experiment $1$. In fact the full SR model had a better AIC.}. 

Next, we compared the remaining models (the winner in each category) using BIC for each experiment, Table~\ref{tab:NLL-bic}.

 \begin{table}[ht]
\caption{Model fit result- BIC comparison} 
\centering 
\begin{tabular}{c c c c c c c c c} 
\hline\hline 
BIC & Baseline & Q-learning & MBRL & MB-RemR & SR & SR-ShPath & CMBRL & Hybrid SR-MB \\ [0.5ex] 
\hline 
Exp $1$ & 2571 & 2266 & - & 1891 & - & \textbf{1518} & 1917 & 1520 \\
Exp $2$ & 2638 & 2479 & 1853 & - & 1896 & - & 1997 & \textbf{1848} \\
Exp $3$ & 2626 & 2451 & 1832 & - & 1809 & - & 2255 & \textbf{1738} \\ [1ex] 
\hline 
\end{tabular}
			\begin{tablenotes}
  				\footnotesize MBRL = Model-based RL;  RemR = Remembers the Last R; ShPath = Find the Shortest Path; SR= Successor Representation; CMBRL = Cubed Model-based RL.
  			\end{tablenotes}
\label{tab:NLL-bic} 
\end{table}

In experiment $1$, the SR model which finds the shortest path (with $3$ free parameters) had the lowest BIC. Note that the hybrid SR-MB model had also a very low BIC ($1520$) although it had $7$ free parameters. In experiments $2$ and $3$, the best model was the hybrid SR-MB. Baseline model, Q-learning and Cubed model-based RL had the worst fit as we expected. The full model-based RL was among the top three models in all experiments. In experiments $1$ and $3$, it was the third best model but in experiments $2$, it held the second place. Note that all of these results were computed by the data in the learning blocks.

One way to compare these models is to treat BIC/AIC as the log model evidence for each participant and investigate if there is a significant difference between the BIC (AIC) for a pair of models, \cite{khodadadi_learning_2017}. However, this needs many pairwise comparisons. Moreover, the results based on AIC and BIC are not consistent to some extent \footnote{Probably because the comparisons are affected more by possible outlier values in BIC or AIC.}: based on AIC, there is much stronger evidence that the hybrid SR-MB model is the best model in experiment $1$. A more sophisticated approach for comparing this number of models is to compare their Exceedance Probability (EP) which have been proposed by Stephan et al., \cite{stephan_bayesian_2009}. In this algorithm, each model was treated as a random variable and the probability of generating (participants') data by a model was defined by a multinomial distribution described by a Dirichlet distribution. Stephan et al. used a variational Bayes method to estimate the Dirichlet distribution's parameters based on marginal likelihood of each model, \cite{stephan_bayesian_2009} \footnote{We used the MATLAB code developed by Samuel J. Gershman \cite{gershman_empirical_2016}.}. Given the parameters of the Dirichlet distribution, it is possible to compute the probability that a model is more likely than any other model \footnote{To compute these probabilities, it is necessary to have an estimate of the marginal likelihood of each model $m$ for the data set $D$, i.e. $p(D|m)$.}. The best model has the highest EP (probability closer to $1$). Table~\ref{tab:EP-bic} shows the EP of six models that we discussed above.

In experiment $1$, the hybrid SR-MB model and the SR-ShPath model were competing with each other and as we expected, the EPs calculated by AIC and BIC were not consistent. In experiment $2$, the full model-based RL was more or less similar to the hybrid SR-MB model while in experiment $3$, the superior fit was provided by the hybrid SR-MB model. These results confirmed our previous findings that the heuristic-based models were not able to explain the data in a more complicated environment. The hybrid SR-MB model was the best model in all three experiments if we only considered the AIC results (and AIC-based EP). For the next section, we chose the best three models in each experiment and compared their predictions in the test blocks.
  
\begin{table}[ht]
\caption{Exceedance probability of different models in $3$ experiments using BIC and AIC.} 
\centering 
\begin{tabular}{c c c c c c c c c c} 
\hline\hline 
EP & Baseline & Q-learning & MBRL & MB-RemR & SR & SR-ShPath & CMBRL & Hybrid SR-MB\\ [0.5ex] 
\hline 
EP- BIC - Exp $1$ & $0$ & $0$ & - & $0$ & - & $0.999$ & 0 & $0.0001$ \\
EP- AIC - Exp $1$ & $0$ & $0$ & - & $0$ & - & $0.247$ & $0$ & $0.753$ \\
EP- BIC - Exp $2$ & $0$ & $0$ & $0.5281$ & - & $0.0003$ & - & $0.0003$ & $0.4711$ \\
EP- AIC - Exp $2$ & $0$ & $0$ & $0.4314$ & - & $0$ & - & $0.0002$ & $0.5683$ \\
EP- BIC - Exp $3$ & $0$ & $0$ & $0.1412$ & - & $0$ & - & $0$ & $0.8588$ \\
EP- AIC - Exp $3$ & $0$ & $0$ & $0.1074$ & - & $0$ & - & $0$ & $0.8926$ \\ [1ex] 
\hline 
\end{tabular}
 	\begin{tablenotes}
  		\footnotesize MBRL = Model-based RL;  RemR = Remembers the Last R; ShPath = Find the Shortest Path; SR= Successor Representation; CMBRL = Cubed Model-based RL. 
  	\end{tablenotes}
\label{tab:EP-bic} 
\end{table}

\subsection{Model Prediction In The Test Blocks}
So far, we used the learning phase data to train our models and then used their quantitative fit to evaluate them. As we presented above, the hybrid SR-MB model was dominantly the best model (among our proposed heuristics and models) in explaining the data in all three experiments using AIC. These results were not so clear when we compare the BIC and the EP of these models. The SR-ShPath and the MBRL had a better fit in experiment $1$ and $2$ respectively. Therefore, we took our model comparison to another level and investigated their predictions in the test phase. Using the knowledge of the training phase, we examined whether these models can find the optimal path in an environment where they have not experienced yet. We evaluated these predictions using two approaches: a) calculate the log-likelihood of each model during the test phase and compare these results (similar to what we have done in the training phase); b) using the best fitted parameters, compare models' prediction at critical cells in the test phase.

 \begin{table}[ht]
 \caption{Exceedance probability of different models in the test blocks.} 
 \centering 
 \begin{tabular}{c c c c c c} 
 \hline\hline 
 EP &  MBRL & MB-RemR & SR & SR-ShPath & Hybrid SR-MB\\ [0.5ex] 
 \hline 
 EP- Exp $1$ & - & $0$ & - & \textbf{0.6775} & $0.3225$ \\
 EP- Exp $2$ & \textbf{0.568} & - & $0.2018$ & - &$0.23$ \\
 EP- Exp $3$ & \textbf{1} & - & $0$ & - & $0$ \\ [1ex] 
 \hline 
 \end{tabular}
  	\begin{tablenotes}
   		\footnotesize EP calculated in the test phase. As in the training blocks, the model with simple heuristic win in experiment $1$ but in experiment $2$ and $3$ the model-based RL which estimates the reward structure and uses that in planning provides a better fit to data. 
   	\end{tablenotes}
 \label{tab:EPTest} 
 \end{table}  
 
Table~\ref{tab:EPTest} summarizes the EP of the three most successful models discussed in model fitting section. The heuristic-based SR model (which finds the shortest path) had the greatest EP in experiment $1$. In experiments $2$ and $3$, the model-based RL was the best model in explaining the test trials' data. To our surprise, the hybrid SR-MB model failed in all of these experiments. One possibility that can explain the failure of hybrid SR-MB model is that in our model fitting, we only used the learning blocks to train these models. When the environment was not changing, even a pure SR model (with or without heuristics) provided a better fit than a model-based RL. But the SR model needs to experience new trajectories when the transition matrix changes. When we linearly combined the SR algorithm with the model-based RL, greater weights were assigned to the SR model's prediction \footnote{The average weights for the SR model in our experiments were close to $1$ and almost $0$ for the model-based RL.}. Therefore, the model-based RL had small or zero contribution in predicting the re-planning trials' data in the test phase.

One solution to this problem is to use a mixture model which switches from the SR model in the planning trials to the model-based RL mechanism in the re-planning trials. We did not fit this model but one possibility to implement this idea is to activate a switch to detect surprising changes of the environment. Then it can signal the hybrid SR-MB model to let the model-based RL take over the decision making process. Assuming that the model switches from training to test can post hoc fit all the data the best, because it uses the best fitting model for training and then switches to the best predicting model for test. However, future research is needed to identify the switching mechanism and design experiments to a priori test this mechanism.

We also compared models' predictions at cell $8$ after participants experienced a random blockage in the optimal path in experiment $1$. In these trials, the optimal behavior is to go up to cell $7$ and then reroute for the second optimal path (path C) to reach the goal. Note that in experiment $1$, $13$ out of $40$ trials in the test blocks were re-planning trials. Also, that the starting position and the goal were fixed at $3$ and $27$. As we expected, the SR model which finds the shortest path (or basically any SR model) was not sensitive to the changes of the transition structure in the environment. The probability of selecting the right action at cell $8$ was still greater than other actions ($.63$), although path B was not available anymore. The predicted probability of selecting the up action, by SR model, was $0.08$. However, in the model-based RL, the probability of selecting the right action is the lowest, $0.03$. Similar patterns were found in experiment $2$ and $3$.\\

\section{Discussion}
Planning in a stochastic environment is challenging. It becomes even more challenging when the environment is unknown to us. No matter how complicated these problems are, we mainly use our previous experiences to deal with them. Sometimes the environment changes and forces us to change or modify our plan. As a result, we update our plan every now and then to make sure our plan becomes a success. In this article, we tried to study real life planning problems in a simplified situation using our grid world experiment. Using this framework, we developed three experiments to investigate planning and re-planning in humans while learning an unknown environment. After $6$ blocks of training in a $4$ by $7$ grid, participants' planning skill was tested. At the beginning of the $7$th block (the pretest block), we warned them that their score will be converted to monetary reward (the exchange rate was $0.01$). The majority of our participants ($19$ out of $19$ in experiment $1$, $30$ out of $36$ in experiment $2$ and $30$ out of $32$ in experiment $3$) were able to find the optimal path. This is in contrast to previous studies that showed people are more likely to be sub-optimal in the description-based decision trees involving a probabilistic reward, \cite{hey_strategies_2011}, \cite{hotaling_dft-d:_2012}. It has been proposed by \cite{erev_anomalies_2017} that people show planning biases in description-based decision trees problems, but not in experience based learning. In what follows, we discuss our findings and contributions, separately.

\subsection{Optimal planning in a grid world}
Simon and Daw \cite{simon_neural_2011} investigated the neural bases of planning in a dynamic maze comparing two well established RL models (TD learning and model-based theory). Participants navigated through a virtual maze (with $16$ rooms) to earn the maximum possible reward. They found that the BOLD signal related to the value and the choice in the striatum is correlated with the prediction of model-based theory \cite{simon_neural_2011}. In their design, participants did \textit{not} need to estimate the expected values of the rewards because rewards were deterministic and the maze, although was constantly (and randomly) changing, was known to participants. Therefore, planning was equivalent to finding the new \textit{shortest} path to reach the goal at each state by relearning the (new) environment while knowing the general structure of the maze. However, in our design, with the stochastic reward structure and deterministic obstacles, the shortest path was not optimal and participants did not need to relearn the configuration. 

The basic configuration of our grid world was similar to the detour problem in \cite{tolman_studies_1946}, but we modified the grid world into a stochastic environment where finding the shortest path was not optimal anymore. It is important to note that our grid world had two distinctive features that encourage goal-directed behavior: first, the (hidden) punishments were probabilistic, and second, the starting and goal positions were randomly located in different cells. In order to find the optimal path, participants needed to learn and compare the expected values of different paths. Employing probabilistic rewards enabled us to represent this problem in a decision tree framework to study (optimal) planning similar to experience-based decision-making tasks, Fig.~\ref{MM01}. 

\subsection{Re-planning in humans}
Through simple error-driven learning rules (e.g. TD learning), the model-free RL selects the action that leads to a greater reward (outcome) more frequently. The model-free RL stores the (action-) state values without learning the model of the system and therefore it is computationally simple. However, when the environment changes similarly to what we have in our grid world, a blockage in the optimal path, these state-action values are useless and need to be learned again! Note that the changes in the environment are not necessarily limited to a transition function, but also could be related to reward (e.g. reward devaluation), \cite{dolan_goals_2013}. This inability to be adapted to varying circumstances makes the model-free RL error-prone when the structure of the task is changing, \cite{daw_uncertainty-based_2005}.

On the other hand, the model-based RL can update the state-action values globally due to its knowledge (or representation) of the system. For instance, in our grid world experiment, when the optimal path is blocked, the model-based RL only changes the transition function for that path. Consequently, this update changes the state-action values (Q-values) using a dynamic programming algorithm. This flexibility enables the model-based RL to pursue the goal without the need to experience and learn the environment again (goal-directed approach). A large amount of research has focused on how these two approaches cooperate or compete with each other in different areas (from clinical studies to neuroscience) using different tasks, \cite{dolan_goals_2013}, \cite{decker_creatures_2016}, \cite{akam_simple_2015}, \cite{boureau_deciding_2015}, \cite{gillan_model-based_2015}, \cite{keramati_speed/accuracy_2011}, \cite{lee_neural_2014}, \cite{schad_processing_2014}, \cite{kool_cost-benefit_nodate}, \cite{skatova_extraversion_2013}, \cite{sebold_model-based_2014}, \cite{momennejad_successor_2016}, \cite{russek_predictive_2017}, \cite{keramati_adaptive_2016}.

When a change occurs in our environment (e.g. grid world), we can not rely on our habits anymore. We need to update our knowledge and modify our original plans to accomplish our goal (re-planning). To the best of our knowledge, re-planning in humans has been rarely studied, \cite{momennejad_successor_2016}, \cite{gershman_successor_2012} with only two-step decisions. We were able to analyze re-planning behavior by randomly blocking the optimal path in our test blocks. In experiment $1$ and $2$, the starting and goal positions were fixed during the test phase, but, in experiment $3$, we randomized these pairs. Because of the stochastic nature of the payoffs, our results are not comparable directly with the rats used by \cite{tolman_cognitive_1948}, but we showed that humans are capable of modifying their plans when there is a change in the environmental circumstances. 
 
\subsection{Computational modeling}
We used $12$ different models to fit the choice data. The baseline model selects a random action at each cell regardless of what participants have experienced. This model was the simplest model in our benchmark. The traditional Q-learning algorithm was not able to explain the data because the starting and goal positions in the environment in our experiments were not fixed. In addition, when a change happens in the environment, the whole set of the action-state values needs to be updated (again by extensive amount of learning and exposure to the new environment). 

The (full) model-based RL tries to learn the model of the environment by estimating the transition and reward functions. This knowledge (of the environment) is later used to generate Q-value for each action. The model-based RL had the best predictions in the test blocks. In model $7$, we restrict the model-based RL in its spatial search. Instead of a complete tree search that is commonly used in value iteration, we confine the model's planning depth to its $k$th nearest neighbors ($k$ is a free parameter). While this modification can decrease the computational costs, for many pairs the best fitted $k$ leads to a full tree search.  

In addition to model-based and model-free RL, there is another alternative, SR, which is more flexible than model-free RL and computationally simpler than model-based RL, \cite{dayan_improving_1993}. SR calculates the state values using both reward and a successor map which stores the expected and discounted future states' occupancies. In case of reward devaluation, SR's behavior is similar to model-based RL but when there is an alteration in the transition structure, it fails to adapt to the change (similar to model-free RL), \cite{gershman_successor_2012}, \cite{kulkarni_deep_2016}, \cite{momennejad_successor_2016}. 

Although the hybrid SR-MB model provided a better account for participants' choices in the learning blocks, it failed to predict the re-planning behavior in the test blocks. Since the candidate models were not trained by the test blocks' data, they needed to \textit{generalize} their knowledge (from the learning blocks) to perform optimally in the re-planning trials when the optimal path was randomly blocked. One solution to this problem is to use a mixture model which switches from the SR model in the planning trials to the model-based RL mechanism in the re-planning trials. It can post hoc fit all the data the best, because it uses the best fitting model for the training blocks and then switches to the best predicting model for the test blocks. 
  
Another possible candidate to explain the data can be a multi-step planning model proposed by Sutton et al., \cite{sutton_between_1999} which introduces the concept of “temporal abstraction” of actions that are interrelated. So instead of selecting an action for each state, a sequence of actions can be chunked and used. This sequence is called an \textit{option} and this framework is called hierarchical reinforcement learning (HRL), \cite{sutton_between_1999}, \cite{botvinick_hierarchically_2009}. Multi-step planning models (MSP) can be implemented in different ways. For instance, a MSP model can update the predetermined plan if any surprising or unexpected event happens (any changes in the transition function or unexpected punishment). So the model modifies the original plan more often at the beginning of learning and less when it builds the representation of the task (with fewer surprises). Also, the model can plan for a couple of steps until it gets to predefined subgoals (like a hub in the grid world) and then updates its estimates before proceeding to the final goal (or next subgoal), \cite{huys_interplay_2015}. The current design did not allow us to specify a unique subgoal for different pairs but it can be a direction for the future research.\\

\subsection{Applying heuristic-based models}
Our heuristic-based models are inspired by some of our participants' data. Avoiding the greatest loss is useful in curtailing the decision tree search, but it is not optimal. It only guarantees not taking the worst action in the cells surrounding the greatest loss, though, it does not provide any policy in other cells. In other words, it is reduced to the baseline model at cells that are not adjacent to cell $11$ and cell $15$ and fails to explain our data. In \cite{huys_bonsai_2012}, they examined whether a big loss at early stages of a decision tree could interfere with finding the optimal path later or not. They found that participants are reluctant to take the paths with the greater loss although it might be counterproductive and they could earn a bigger reward later (on that path). In our experiment, we did not ask this question directly, but we found that there is a great avoidance toward the great loss of $-75$ \footnote{The hot stove effect \cite{denrell_adaptation_2001} suggests that avoiding bad outcomes can form information asymmetry between available options. We randomized the starting points and the goals to make sure that participants have unbiased exposure to the environment.}. Also in the re-planning trials, there were only a few participants (three out of eighty seven in all $3$ experiments) who chose the path with the greatest punishment. We should clarify that in our experiments, participants received $100$ points at the goal position regardless of their chosen path. Hence, avoiding large losses in our experiments could not be compared with pruning behavior in \cite{huys_bonsai_2012}. 

One way to differentiate between the cells with regular punishments and the cells with great losses is to use different discount parameters to weight future and immediate rewards differently as it was suggested in \cite{huys_bonsai_2012}. Instead, in model $3$, we used different rates (e.g. $\alpha_{1_{HS}}$ for the great loss of $-75$) to estimate the reward at these cells. One of the models in this category assigns one special rate, $\alpha_{s1}$ (and possibly one special forgetting factor, $\alpha_{s2}$) for estimating the reward at cell $11$ for a loss of $-75$ while treating the other cells, with regular or non-regular losses, equally. The model which captures participants' tendency to avoid large losses with only two rates is inferior to our proposed model with three distinctive rates for estimating reward function.

There are few participants who chose the shortest path. Because the rewards are probabilistic, this approach can be the worst strategy to apply. If path A1A2 is involved in catching the goal in a pair, it will be the shortest path but costs the greatest loss. This is not consistent with the observed pattern in the majority of our participants' data. 

Our final simple heuristic-based model (which was more successful than the others) remembers the last punishment at each cell. Before experiencing the non-regular losses at critical cells ($11$, $9$, $17$, $16$ and $19$ in experiment $2$ and $3$; $15$, $16$ and $21$ in experiment $1$), the model can not prioritize available paths optimally. But with one (loss) experience, it starts to select actions that lead to  greater expected rewards. Remembering the last experience simplifies the reward estimation procedure, but it also promotes wrong predictions after experiencing a regular punishment at critical cells. Thus, this model fails except for the simple case of experiment $1$.\\
 
\subsection{Related research}
Unpacking different aspects of planning problems has been an interest to many researchers in different fields. While economic theory provides a normative approach for these problems, results from psychological experiments showed that people used simple heuristics, which were not necessarily optimal to maximize their payoffs in a multi-stage decision tree problem, \cite{hey_how_2007}, \cite{hey_strategies_2011}, \cite{hotaling_dft-d:_2012}. In these experiments, the probability of transition at each choice node was also known to the subjects. They found that, except for a few people who planned ahead (e.g. rational planners), the majority of participants were no-planners or planners with mixed heuristics, \cite{hey_strategies_2011} and \cite{hey_people_2005}. The latter group tried to simplify the problem by using less information or mixing different strategies (i.e. a combination of local optimization at one level and random guessing at a different level) whereas the rational planners used computations similar to backward induction \cite{degroot_optimal_2005}. Yet, some of the strategies mentioned above had certain predictions that were not consistent with the experimental data. For instance, one strong prediction in backward induction is dynamic consistency \footnote{Dynamic consistency assumes that decision makers follow the original plan that they made for their future choices. In \cite{johnson_multiple-stage_2001}, they found that people usually change their choices as they move forward in a decision tree (rather than following the original plan) and this inconsistency increases when they deal with longer decision trees.}, which has been questioned before, \cite{busemeyer_dynamic_2000}, \cite{johnson_multiple-stage_2001}. \cite{hotaling_dft-d:_2012} tried to address these concerns by proposing a new model, the Decision Field Theory-Dynamic model (DFT-D), which was able to distinguish the aforementioned patterns of participants' data and also provided a dynamic account of the decision making process at decision nodes. DFT (among with other sequential sampling models) assumes that the subject accumulates noisy information favoring each choice alternative until the evidence favoring one of the alternatives meets a decision threshold, \cite{hotaling_dft-d:_2012}, \cite{khodadadi_learning_2014}. The DFT-D model is a cognitive-dynamical model that extends DFT for multistage decisions (in planning).

Using decision trees for studying sequential choices has several drawbacks. First, when the number of decision nodes increases (in bigger decision trees), participants find it too difficult to consider all the potential future consequences (of each decision) and thus might start choosing at random or adopt simple heuristics \cite{hey_people_2005}, \cite{huys_bonsai_2012}. Thus, it remains unclear if humans' sub-optimal performance in these tasks is a reflection of their planning behavior or their poor understanding of the environment with which they are dealing. Second, the traditional design of decision trees restricts our design to fix the starting position at the top of the tree; Doing otherwise would result in planning problems being reduced to a single one-time decision. For instance, in a $2$-step decision tree, the starting point cannot be placed in the second layer. There are few studies that used $3$-step decision trees; however, they fixed the starting position at the top of the decision tree and allowed participants to use Notepad to write comments and remarks \cite{hey_strategies_2011}. 

To address these issues, many researchers started to use spatial framework because it is more natural for participants for planning and multi-step decision making rather than a decision tree for which they have little or no experience in real life. In our experiment, we minimized spatial reasoning by displaying the environment on the screen along with the participants' position, the goal and all the feasible paths between them. More importantly, unlike \cite{yoshida_resolution_2006}, \cite{gallistel_computations_1996}, \cite{doeller_parallel_2008}, we \textit{did not} manipulate viewpoints, locational uncertainty and external cues. Therefore, only decision making and learning theories were directly relevant to our problem. 

\subsection{Conclusion}
There are many theoretical studies in reinforcement learning on how to train an agent in an unknown, complicated environment by focusing on reducing the computational costs and optimizing the search algorithms \cite{koenig_improved_2002}, \cite{ersson_path_2001}, \cite{thrun_learning_1998}, \cite{simmons_probabilistic_1995}, \cite{fakhari_quantum_2013}, \cite{meyer_map-based_2003}, \cite{barto_recent_2003}, \cite{doya_reinforcement_2000}, \cite{mataric_reinforcement_1997}, \cite{tani_model-based_1996}. Few of them, however, have been tested with behavioral data in sequential-choice tasks, \cite{walsh_navigating_2014}. In this study, we were able to simplify one of the real life situations (navigation between two places) in our grid world experiment and evaluated predictions of the reinforcement learning theories with respect to choice data. Our design integrated experience-based decision-making into a classical decision tree problem. We showed that people are capable of revising their plans when an unexpected event occurs and that optimal re-planning requires learning the model of the environment (as in \cite{tolman_studies_1946}). 
  
\section{References}
\bibliography{references}

\end{document}